\documentclass[letterpaper, 10 pt]{article}  
\usepackage[top=2in, bottom=1.5in, left=1.5in, right=1.5in]{geometry}

\usepackage{amsmath}
\usepackage{amsfonts}
\usepackage{amsthm}  

\usepackage{graphicx} 

\usepackage{algorithm}
\usepackage{algorithmic}

\usepackage{amsthm}

\usepackage{hyperref}

\usepackage{bm}
\usepackage{amsbsy}
\usepackage[usenames,dvipsnames]{color}
\usepackage{accents}

\newcommand{\hide}[1]{}
\newcommand{\dnote}[1]{}

\newcommand{\spt}[1]{\mathcal{#1}}
\newcommand{\mat}[1]{{\boldsymbol{#1}}}

\newcommand{\f}{\mat{f}}

\newcommand{\g}{\mat{g}}

\newcommand{\h}{\mat{h}}
\newcommand{\w}{\mat{w}}
\newcommand{\q}{\mat{q}}
\newcommand{\qd}{{\dot{\mat{q}}}}
\newcommand{\qdd}{{\ddot{\mat{q}}}}

\newcommand{\xdd}{\ddot{\mat{x}}}

\newcommand{\mv}{\mat{v}}

\newcommand{\A}{\mat{A}}

\newcommand{\I}{\mat{I}}
\newcommand{\J}{\mat{J}}

\newcommand{\M}{\mat{M}}

\newcommand{\zero}{\mat{0}}

\newcommand{\torq}{\mat{\tau}}

\newcommand{\wt}[1]{\widetilde{#1}}

\newcommand{\flearn}{\f_{\text{offset}}}
\newcommand{\fid}{\f_\text{id}}

\title{%
  \bf DOOMED: \\
  {\Large Direct Online Optimization of Modeling Errors in Dynamics}
}

\author{Nathan Ratliff$^{1}$, Franziska Meier$^{2,3}$, Daniel Kappler$^{1,2}$ and Stefan Schaal$^{2,3}$\\
\small
$^{1}$ Lula Robotics Inc.\\
\small
$^{2}$ Autonomous Motion Department, MPI for Intelligent Systems, T\"ubingen, Germany\\
\small
$^{3}$ CLMC Lab, University of Southern California, Los Angeles, USA\\
}
\date{\small \today}

\begin{document}

\maketitle

\newcommand{\wideM}{\ensuremath{\widehat{\M}}}
\newcommand{\wideh}{\ensuremath{\widehat{\h}}}
\newcommand{\wtorq}{\ensuremath{\widehat{\torq}}}

\begin{abstract}
  It has long been hoped that model-based control will improve tracking
  performance while maintaining or increasing compliance.  This hope hinges on
  having or being able to estimate an accurate inverse dynamics model. As a
  result, substantial effort has gone into modeling and estimating dynamics
  (error) models. Most recent research has focused on learning the true inverse
  dynamics using data points mapping observed accelerations to the torques used
  to generate them.  Unfortunately, if the initial tracking error is bad, such
  learning processes may train substantially off-distribution to predict well
  on actual \textit{observed} acceleration rather then the desired
  accelerations. This work takes a different approach. 
  We define a class of gradient-based online learning algorithms we
  term Direct Online Optimization for Modeling Errors in Dynamics (DOOMED) that
  directly minimize an objective measuring the divergence between
  actual and desired accelerations. Our objective is defined in terms
  of the true system's unknown dynamics and is therefore impossible to
  evaluate. However, we show that its gradient is measurable online
  from system data. We develop a novel adaptive control approach based
  on running online learning to directly correct (inverse) dynamics
  errors in real time using the data stream from the robot to
  accurately achieve desired accelerations during execution.
\end{abstract}

\section{Introduction and Motivation}
Acceleration policies are natural representations of motion: how
should the robot accelerate if it finds itself in a given
configuration moving with a particular velocity? Writing down these
policies is easy, especially for manipulation platforms with
invertible dynamics. And, importantly, it is very easy to shape these
policies as desired. Simulation is as easy as simple kinematic forward
integration, which means that motion optimizers, especially those
generating second-order approximations to the problems (LQRs
\cite{OptimalControlEstimationStengel94}), can analyze in advance what
following these policies means. Additionally, LQRs can effectively
shape a whole (infinite) bundle of integral curve trajectories
simultaneously by adjusting (maybe through a training process) the
strengths and shape of various cost terms. DMPs \cite{Ijspeert2013}
are another form of acceleration policy, which can be shaped through
imitation learning for instance.
Unfortunately, in practice following acceleration policies is hard.
The fidelity of the dynamics model defines how accurately we can
simulate the behavior of the robot, and it's widely understood that we
can never actually simulate the robot well enough to sufficiently
predict it's behavior on the real system. Additionally,
tracking a commanded acceleration on the real system is hindered by
inaccuracies of approximate dynamics models.

Thus, we typically resort to error feedback control. The acceleration
policies are integrated to obtain position and velocities that can be
tracked. Given these, the control law typically is a combination of
the predicted inverse dynamics torque plus a PID term on the state
error. This control approach works well in practice
\cite{Righetti_AR_2013}, however not having to count on the feedback
term to reject modeling errors could increase compliance further.
Typically, the approximate dynamics model might not work equally well
throughout the state space of the system - for instance it may be
tuned to be a better approximation on slower movements. In this
scenario, to also track accurately faster movements, we would have to
tune feedback terms to be able to do so. This would typically mean
that we need to use higher feedback gains than wished, resulting in a
reduction in compliance.
Thus, there has been an effort in bringing the promise of machine
learning to the world of model-based control to estimate better
inverse dynamics (error) models \cite{Vijayakumar2000,
  Nguyen-Tuong2008a, Gijsberts2013, Meier2014a}. These recent approaches
exclusively consider minimizing the loss between the applied torque
and the models predicted torque at the actual state and acceleration.
We term these methods \emph{indirect loss minimization} methods in
this work. While, intuitively, they learn the true inverse dynamics
model and should thus be exactly what we want to estimate, there are
some issues when using this indirect loss. In particular, they train
on slightly off-distribution in the sense that they train to predict 
well on \emph{actual} observed accelerations rather than 
the \emph{desired} accelerations .

A simple thought experiment on controlling around stictions on a
real system helps to illustrate the main issue, when learning an error
model using the indirect loss. When experiencing stiction, you command
the system to accelerate at a desired rate, estimate the required
torques for this and send them, but nothing happens, meaning the
system stays in its current state. Now, you would want to send a
different amount of torque to actually achieve these desired
accelerations at the next time step. But the model's only data
at this point is the applied torque and associated zero acceleration.
The learner will update to map an acceleration of zero to this applied
torque, but any improvement to the prediction at the actual desired
acceleration is residual and model-class dependent at this point. In
some cases, the system may never learn to produce the torque needed to
break through the stiction since it focuses only on updating the
predicted torque for a zero desired acceleration.
Moreover, there is not
actually a single ``correct'' value that the function should be training toward
in this scenario
since the dynamics
is not actually invertible in the presence of unknown stiction.
So in this particular case, the traditional problem is technically ill-posed.

In this work, we propose instead to explicitly minimize the error
between the desired and actual accelerations using online learning
tools. The objective we define is based on the true system's
\textit{unknown} dynamics and is therefore impossible to evaluate.
However, we show that its gradient is \emph{measurable}\footnote{
  We use ``measurable'' in the broad sense including
  quantities estimated from measurable quantities
  to be consistent in our terminology between cases when we can directly 
  measure a quantity (accelerations via accelerometers, for instance) and 
  cases when we have to derive the result through a statistical estimator
  (accelerations via finite-differencing). Note that depending on 
  the estimator used, there may be differences in statistical variance
  of the estimate. Section~\ref{sec:TelescopingFiniteDifferences} 
  discusses variance more carefully
  and shows why common estimators for these gradients are often usable in 
  practice despite the noise in the raw estimates.
} online
system data enabling the application of numerous online learning
approaches. In contrast to the above \emph{indirect loss} approach,
this \textit{direct loss} minimization approach is both well-posed and
able to leverage a long lineage of well-studied online learning
methods to adapt quickly and achieve good tracking in this real time
setting.
This algorithm leverages the data streaming from the robot to correct
the inverse dynamics model on the fly by tuning the a correction model
until it 
achieves the right \emph{acceleration}. This enables accurate and direct
execution of raw acceleration policies acting on state feedback,
without requiring the purely feedforward (and therefore unreactive)
forward integration of a state-feedback-independent trajectory
tracking signal (which is often used in practice).

We start out by reviewing the various backgrounds of inverse dynamics control, adaptive control and online gradient descent in Section~\ref{sec:background}. Then, in Section~\ref{sec:algorithmDerivation}, we derive our proposed approach \emph{Direct Online Optimization of Modeling Errors in Dynamics}. In Section ~\ref{sec:experiments} we extensively evaluate our work both in simulation and on a real robotic system. Finally, after concluding in Section~\ref{sec:conclusion} we present some interesting theoretical connections between our approach and PID-control in Section~\ref{sec:theoretical_connections}.

\section{Background}
\label{sec:background}
Our work has connections to a variety of research areas such as
inverse dynamics control, adaptive control, and the use of modern
machine learning techniques such as online gradient descent. In the
following we, provide a (non-exhaustive) review and introduction into
these topics.
\subsection{Inverse Dynamics}
The dynamics of any classical dynamical system \cite{ClassicalMechanicsTaylor} 
can be expressed as
\begin{align} \label{eqn:dynamicsGeneralForm}
  \torq &= \M(\q)\qdd + \h(\q, \qd),
\end{align}
as derived from the Principle of Least Action.
$\M(\q)$ represents the generalized inertia matrix and $\h(\q, \qd)$ collects
the modeled forces including gravitational, Coriolis, centrifugal
forces, and viscous and Coulomb friction. 
Model-based control \cite{An:1988wi, craig2005introduction} constructs
a model of these dynamics to predict the torques $\torq$
required to realize desired accelerations $\qdd$ in the current state
$\q, \qd$ by estimating the true dynamical functions $\M(\q)$ and
$\h(\q, \qd)$ using data. We generally denote these estimated
parameters as $\wideM$ and $\wideh$. In particular, for manipulators,
rigid-body assumptions commonly substantially simplify the mathematics
of the estimation problem. These Rigid Body Dynamics (RBD) models are
linear in the unknown model parameters, and permit the use of standard
linear regression techniques to estimate $\wideM$ and $\wideh$:
\begin{align} \label{eqn:RBDlinearform}
  \torq &= \wideM(\q)\qdd + \wideh(\q, \qd) = \mathbf{Y}(\q, \qd, \qdd) \mathbf{a},
\end{align}
for the appropriate functions $\mathbf{Y}$, as has been shown in \cite{An:1985el}.

Equation~\ref{eqn:dynamicsGeneralForm} is written in \textit{inverse
  dynamics} form. Given an acceleration $\qdd$, it tells us what
torque $\torq$ would produce that acceleration. Inverting the
expression gives the generic equation for forward dynamics:
\begin{align} \label{eqn:forwardDynamicsGeneralForm}
  \qdd = \M^{-1}\big(\torq - \h\big).
\end{align}
This equation expresses the kinematic effect of applied torques
$\torq$ in terms of the accelerations $\qdd$ they generate.

\subsection{Online Learning of Inverse Dynamics}
While, inverse dynamics control, with an estimated RBD model is
successfully deployed on modern manipulation platforms
\cite{Righetti_AR_2013}, the inaccuracies of the estimated dynamics
model remains an open issue. The better we can model and
predict the dynamics, the less we rely on error feedback control to
account for modeling errors. Thus, the learning of inverse dynamics
(error) models is an active research area.
The problem of inverse dynamics learning has many facets, and is
tackled from many different fronts. A focus of recent research progress has
been scaling up modern function approximators so that real-time
(online) model learning becomes feasible \cite{Gijsberts2013,
Vijayakumar2000, Meier2014a, Nguyen-Tuong2008a}. These methods
attempt to learn a \emph{global} inverse dynamics model, that can be updated
online and used for real-time prediction. The models retain a memory
and theoretically improve with repeated execution of similar tasks,
and as such are often categorized as \emph{learning control} methods.
Computational efficiency and robustness has been the focus of this
research path.
Our proposed approach, falls into the category of \emph{adaptive
  control} \cite{AdaptiveControl2008}. Adaptive control approaches
update either controller or system model parameters online. As opposed to \emph{learning control} approaches there is no notion of improving over multiple task trials. 
Within this field, a
popular approach has been to utilize the Rigid Body Dynamics (RBD) model to
derive updates for
adaptive control laws \cite{Slotine:1989fd, Craig:1987dc}. For this,
the linear relationship of the RBD parameters and the
torques equation~\ref{eqn:RBDlinearform} is used also for modeling the RBD model error. Then Lyaponuv based update rules are derived to
continuously (in real-time) update the error model.
In contrast, in our work we 
derive gradient-based online learning algorithms for tuning dynamics model
function approximators to directly minimize
the discrepancy between desired and actual
accelerations, and demonstrate their real-world efficacy for 
the case of adapting the torque offset needed to achieve a desired
acceleration. 

\subsection{Adaptive Control on Direct vs Indirect Loss}
In the context of adaptive control, we make the distinction
between \emph{direct} and \emph{indirect} loss minimization approaches.
To explain the difference, we first take a more detailed look at the
error made when using an approximate RBD model.

For any given desired acceleration $\qdd_d$, we calculate torques
$\torq = \wideM\qdd_d + \wideh$ using our approximate inverse dynamics
model, but when we apply those torques they are pushed through
the \emph{true} system which may differ substantially from the estimated model:
\begin{align}
  \qdd_a = \M^{-1}\Big(\big(\wideM\qdd_d + \wideh\big) - \h\Big),
\end{align}
where here $\qdd_a$ denotes the actual observed accelerations of the
true system. We emphasize that this true system model is generic and
holds for any Lagrangian mechanical system, without requiring rigid
body assumptions. These true dynamics are typically unknown, so this
expression, thus far, is of only theoretical interest. We will see
below that expressing the structure in this way enables the derivation
of a practical algorithm that we can use in practice.

Rigid body assumptions are often restrictive, introducing a substantial offset between
the true model and the estimated model. We propose, therefore, to learn an 
offset function
$\flearn(\q, \qd, \qdd_d, \w)$ that models the error made by
this approximate RBD model
\begin{align} \label{eqn:ActualAccel}
  \qdd_a = \M^{-1}\Big(\big(\wideM\qdd_d + \wideh + \flearn(\q, \qd, \qdd_d, \w)\big) - \h\Big),
\end{align}
In an extreme case, when the approximate dynamics model is the zero function,
this $\flearn$ function represents the full inverse dynamics model,
which must be trained from data. Often, though, we can take it to be a model
representing only the difference between the true inverse dynamics and the
modeled inverse dynamics (e.g. calculated under standard rigid body
assumptions).

Depending on what type of loss function is used to adapt parameters
$\w$ we distinguish between adaptive control as \emph{direct} and
\emph{indirect} loss minimization approaches. More concretely,
consider an objective of the form:
\begin{align}
  l_\text{indirect}(\w) = \big\|\torq - \f(\q, \qd, \qdd, \w)\big\|^2,
\end{align}
where $f$ is any class of function approximators parameterized by $\w$
predicting the inverse dynamics. In this case, $\torq$ is the actual
applied torques, and $\qdd$ is the corresponding actual accelerations
that were observed.
Adapting parameters based on this loss function attempts to make the
model's predicted torque on the \textit{actual} observed acceleration
more accurate. But in reality, we wanted to achieve the desired
accelerations. This discrepancy is especially problematic when
stictions are involved: all actual observed accelerations are zero
when we are not applying enough torque, thus the error model never
receives data to accurately predict an offset for desired torques.\footnote{
  Depending on the rigidity of the hypothesis class representing the offset
  function, the system may go through an online iterative improvement process
  that tangentially coaxes the predicted torque at the desired acceleration
  to increase over time until it's large enough to push through stiction.
  Basically, predicting a torque $\torq$ at $\qdd_a = \zero$ may induces 
  a slightly larger torque prediction $\torq + \Delta\torq$ at the desired
  acceleration $\qdd_d$. If that's the case, the next training step will
  update the model to predict the slightly larger $\torq + \Delta\torq$
  at $\qdd_a$, which will in turn induce an even larger prediction 
  $\torq + 2\Delta\torq$ at $\qdd_d$, and so on an so forth.
  The process may ultimately converge toward large enough torques
  to break through the stiction, but it is hard to analyze 
  and the property need not hold in general.
}
In these cases, without a state error feedback control term, pushing
through stictions is problematic.

In this work, we instead develop a new adaptive control methodology that
\emph{directly minimizes} the acceleration error:
\begin{align}
  l_\text{direct}(\w) = \big\|\qdd_d - \qdd_a(\w) \big\|_{\M}^2.
\end{align}
In Section~\ref{sec:algorithmDerivation}, 
we derive our approach as a gradient-based
online learning technique,
enabling us to leverage a broad collection of practical and theoretically
sound tools developed by the machine learning community.

\subsection{Online and Stochastic Gradient Descent}

Machine learning often frames the learning problem as one where, given data,
the task is to estimate a model that generalizes that data well (where these
terms are made rigorous in various theoretical settings). But it's often useful
to analyze learning algorithms instead within a setting where data is presented
only incrementally, in the extremely only one data point at a time. This is the
subject of \textit{online learning} (see \cite{Cesa-BianchiPLG2006}).

Online learning, over the past decade, has become a general theoretical 
framework wherein both online learning processes and batch learning processes 
can be analyzed building from a framework called \textit{regret analysis}.
Regret bounds were first studied in the context of online gradient descent
in \cite{zinkevich2003online}, where regret is defined in terms of how 
well the algorithm does on the stream of objectives 
relative do the best it could have done if
it had seen all of the objective functions in advance.
Characteristic of this approach is the lack of assumptions made on the sequence
of objective functions seen. Rather than assuming the data points are 
independent and identically distributed (iid) as is frequently the case
in statistics and machine learning \cite{MachineLearningBishop2007, WassermanAllOfStatistics2010},
these approaches allow the data stream to be anything. The theoretical 
performance of an algorithm is rated purely relative to the best it could 
do. If the sequence is inherently bad, that's ok---the algorithm does as 
well as possible given the difficult problem. If the sequence is good,
then the regret bounds show that the algorithms perform well.

In particular, regret bounds in the online setting can be specialized to give
generalization guarantees if additional assumptions are made on the data
generation process. For instance, if we additionally assume that the data is
iid, then it's possible to produce novel and out-of-sample generalization
bounds that are competitive with (or superior to) the best known guarantees
\cite{GeneralizabilityOnlineLearning2004, SubgradientMMSCRatliff2007}.

This framework is closely related to another widely studied class of algorithms
known as stochastic gradient descent \cite{StochasticLearningBottou2004}, 
which is perhaps the
most commonly used underlying optimization technique in modern large-scale
machine learning \cite{OptimizationLargeScaleML2016}.
Characteristic of stochastic gradient methods is the assumption of noise in the
gradient estimates due to both noise in the underlying data and seeing only
part of the problem (a statistical ``mini-batch'' sample) at any given time.

These techniques are extremely important to our settings, since in online
control we see only streams of data as they're generated by the robot, and, as
we'll see below, our gradient estimates are noisy. Deriving these algorithms as
gradient-based online learning enables us to inherit all of the analytical
techniques and stability properties of these optimizers, as well as a
collection of strongly experimentally verified algorithmic variants designed to
address the real-world problems that arise from large-scale machine learning in
practice \cite{adam,adagrad}.

\section{Direct Online Optimization of Modeling Errors in Dynamics}
\label{sec:algorithmDerivation}

This section derives the most basic variant of our Direct Online Optimization
of Modeling Errors in Dynamics (DOOMED) algorithm for tuning an offset to the 
dynamics model to minimize acceleration errors. We first derive the objective function,
then show how its gradient can be estimated from data.

Equation~\ref{eqn:ActualAccel} expresses the true observed acceleration
achieved on the physical system as a function of the offset function's
parameter setting $\w$. In full this expression takes the form
\begin{align}
  \qdd_{a}(\q, \qd, \qdd_d, \w) &= 
  \M(\q)^{-1}\left[\left(\widehat{\M}(\q)\qdd_d + \widehat{\h}(\q, \qd) + \flearn(\q, \qd, \qdd_d, \w)\right) - \h(\q, \qd)\right],
\end{align}
but we often use the shorthand
$\qdd_a(\w) = \M^{-1}\Big[\big(\fid +
\flearn(\w)\big) - \h\Big]$ for brevity to suppress the
dependence on $\q$, $\qd$, and $\qdd_d$ and emphasize the dependencies
on $\w$. $\fid$ here denotes the modeled approximate inverse
dynamics function. The observed accelerations $\qdd_a$ are then a
function of $\w$.
Now that we have an expression for $\qdd_a(\w)$ for the true
accelerations as a function of the offset function's parameters $\w$,
we can write out an explicit loss function measuring the error between
the desired accelerations and actual accelerations:
\begin{align}
  l(\w) = \frac{1}{2}\|\qdd_d - \qdd_a(\w)\|_{\M}^2.
\end{align}
This error is a common metric most famously used in Gauss's Principle
of Least Constraint.

Note that $\M$ in this expression is the \textit{true} mass matrix,
which we don't know, as is the implicit $\h$ in $\qdd_a(\w)$, so we
can't evaluate the objective directly. However, the Jacobian of
$\qdd_a(\w)$ is
\begin{align}
  \frac{\partial}{\partial \w} \qdd_a(\w) 
  = \M^{-1}\J_{f},
\end{align}
where $\J_f = \frac{\partial \flearn}{\partial \w}$ is the
Jacobian of the offset function approximator. So if we evaluate the
gradient of the expression, we see that the unknown elements all
vanish from the expression:
\begin{align}
  \nabla_\w l(\w) &= \nabla_\w \frac{1}{2}\|\qdd_d - \qdd_{a}(\w)\|_{\M}^2 \\
  &= -\left[\frac{\partial\qdd_a(\w)}{\partial \w}\right]^T\M\big(\qdd_d - \qdd_{a}(\w)\big) \\
  &= -\J^T_f\M^{-1}\M\big(\qdd_d - \qdd_{a}(\w)\big) \\ \label{eqn:basicErrorGradient}
  &= -\J^T_f\big(\qdd_d - \qdd_{a}(\w)\big),
\end{align}
leaving us with a combination of quantities that we can either
evaluate or measure in practice.\footnote{In practice, we measure accelerations
in our experiments using finite-differencing. See 
Section~\ref{sec:RealWorldExperiments} for real-world experiments using noisy acceleration
estimates, and 
Sections~\ref{sec:NoisyAccelerationMeasurements} and \ref{sec:TelescopingFiniteDifferences}
for discussions of how estimator variance affects these algorithms.} 
Interestingly, this expression is
intuitive. The gradient is simply the acceleration error pushed
through the Jacobian of the offset function. We know the desired
acceleration $\qdd_d$, we can easily obtain the true acceleration
$\qdd_a$, and we assume we can evaluate the Jacobian of the offset
function approximator $\flearn$. So despite being unable to
evaluate the objective or fully evaluate the gradient, we can still
obtain the gradient from the running system (online) in practice.
For the experiments in this paper, we use an extremely simple offset
function of the form $\flearn(\w) = \w$ representing the
excess instantaneous torque needed to accelerate as
desired.\footnote{Note that this function is constant, but in
  practice, the online learning algorithm is able to \textit{track}
  the needed offset as it changes across a single movement as a
  function of the state-dependent inaccuracies.} For this simple
offset expression, the Jacobian is just the identity matrix
$\frac{\partial }{\partial \w}\flearn = \I$, so the gradient
is simply the acceleration error.

Note also that if our parameters are hypothesized forces in any task
space, or multiple task spaces, the resulting gradient expression is
again intuitive. Let
$\flearn = \sum_{i=1}^k\J_i^{T}\lambda_i$, where $\J_i$ is
the Jacobian of the task map (e.g. the Jacobian of the end-effector
when the task space is the end-effector space), and $\lambda_i$ is a
hypothesized force applied in the task space (e.g. at the
end-effector). Then $\J = [\J_1^T,\dots,\J^T_k]$, and the gradient of
the loss becomes
\begin{align}
  \nabla_\w l = -\left[
    \begin{array}{c}
      J_1\\
      \vdots \\
      J_k
    \end{array}
  \right]\Big(\qdd_d - \qdd_a\Big)
  =
    -
    \left[
    \begin{array}{c}
      \xdd_1^d - \xdd_1^a \\
      \vdots \\
      \xdd_k^d - \xdd_k^a
    \end{array}
  \right],
\end{align}
where $\xdd_i^d = J_i\qdd_d$ is the desired acceleration through the
$i$th task space, and $\xdd_i^a = \J_i\qdd_a$ is the actual
acceleration through the $i$th task space. In other words, the same
intuitive rule for measuring the gradient holds within any task space:
the gradient is simply the acceleration error as measured in the task
space.

Using this loss function as the risk term in an online regularized
risk objective of the form
\begin{align}
  \spt{L}(\w) = \frac{1}{2}\|\qdd_d - \qdd_{a}(\w)\|_{\M}^2 + \frac{\lambda}{2}\|\w\|^2,
\end{align}
we can write out a simple gradient descent online learning algorithm as
\begin{align}\label{eq:gradient_descent_step}
  \w^{t+1} = (1-\eta^t\lambda)\w^t + \eta^t\J_f^T\big(\qdd_d^{t} - \qdd_a^{t}\big).
\end{align}
Where $\J_f = \I$ (or is the the Jacobian of a mapping to some task space),
this expression is essentially an integral term on the acceleration error with
a forgetting factor. But vanilla gradient descent is the simplest 
online gradient-based algorithm we
could use. By deriving the method as online learning, we now understand
theoretically how to apply an entire arsenal of new, more powerful and
adaptive, online gradient-based algorithms to this same problem to improve
performance.

Next we review two tricks pulled from the combined online learning
literature (or more general stochastic gradient descent machine learning
literature) and the adaptive control literature to remove the potential for
parameter oscillations and track changes in modeling errors while
simultaneously enabling high accuracy for precise meticulous movements.

\subsection{Parameter oscillations in online learning and their
  physical manifestation} \label{sec:onlineLearning}

Parameter oscillations in neural networks are a problem resulting from
ill-conditioning of the objective. The objective, as seen from the the
parameter space (under Euclidean geometry, which is a common easy choice), is
quite elongated, meaning that it's extremely stretched with a highly diverse
Hessian Eigenspectrum.  Gradient descent alone in those settings undergoes
severe oscillations making it's progress slow.  More importantly, in our case,
oscillations resulting from the ill-conditioning of the problem manifest
physically as oscillations in the controller when the step size is too large.
Fortunately, the learning community has a number of tricks to prevent these
oscillations and promote fast convergence.

The most commonly used method for preventing oscillations is the use of momentum.
Denoting $\g^t$ as the gradient at time $t$, the momentum update is
\begin{align}
  \mat{u}^{t+1} &= \gamma \mat{u}^t + (1-\gamma)\big(-\g^t\big) \\
  \w^{t+1} &= \w^t + \eta \mat{u}^{t+1}.
\end{align}
effectively, we treat the parameter $\w$ as the location of 
a particle with mass and treat the objective
as a potential field generating forces on the particle. 
The amount of  mass affects its perturbation 
response to the force field, so larger mass results in smoother motion.

It can be shown that this update can be written equivalently
as an exponential smoother:
\begin{align}
  \mat{u}^{t+1} &= \mat{u}^t  - \eta \g^t, \\
  \w^{t+1} &= \gamma \w^t + (1-\gamma) \mat{u}^{t+1}.\label{eq:exponential_smoothing}
\end{align}
Note that $\g^t$ is still being evaluated at $\w^t$, so it's not exactly equivalent
to running simply a smoother on gradient descent (gradients in our case come from 
evaluations at smoothed points), but it's similar.

This latter interpretation is nice because it shows that we're taking the gradients
and 1. literally smoothing them over time, and 2. effectively operating on a slower 
time scale. That second point is important: this technique works because the 
time scale of the changing system across motions generated by the 
acceleration policies is fundamentally slower than that of the controller. This enables
the controller (online learning) to use hundreds or even thousands of examples
to adjust to new changes as it moves between different areas of the configuration space.

Another, common trick found in the machine learning literature (especially
recently due to it's utility in deep learning training), is to scale
the space by the observed variance of the error signal. When the error signal
has high variance in a given dimension, the length scale of variation is
smaller (small perturbations result in large changes). In that case, the step
size should decrease. Similarly, when the observed variance is small, we can
increase the step size to some maximal value.  In our case, we care primarily
about variance in the actual accelerations $\qdd_a$ (which measures the
baseline noise, too, in the estimates) since we can assume the desired
acceleration $\qdd_d$ signal is changing only slowly relative to the 1ms
control loop. Denoting this $\qdd_a$ variance estimate as $\mv^t$ we 
scale the update as
\begin{align}\label{eq:variance_scaling}
  \mat{u}^{t+1} = \mat{u}^t - \big(\I + \alpha\ \mathrm{\bf diag}(\mv^t)\big)^{-1} \g^t.
\end{align}
Note that this is equivalent to using an estimated metric or Hessian
approximation of the form $\A = \I + \alpha\ \mathrm{\bf diag}(\mv^t)$.

This combination of exponential smoothing (momentum) and a space metric built
on an estimate of variance results in a smoothly changing $\w$ that's still able
to track changes in the dynamics model errors.

\subsection{Adaptive tuning of the forgetting factor}

Indirect approaches to adaptive control (essentially online regressions of
linear dynamics models) often tune their forgetting factor based on the
magnitude of the error they're seeing \cite{RobustAdaptiveControlIoannou2012}.
Larger errors mean that the previous model is bad and we should forget it fast
to best leverage the latest data.  Smaller errors mean that we're doing pretty
well, and we should use as much data as possible to fully converge to zero
tracking error. Adaptively tuning the forgetting factor, which manifests as
adaptive tuning of the regularizer in our case, enables fast response to new
modeling errors while simultaneously promoting accurate convergence to
meticulous manipulation targets. 

The forgetting factor, as describe in detail in
Section~\ref{sec:TelescopingFiniteDifferences}, is the regularization constant
$\lambda$. In our experiments, we utilize algorithmic variants that adapt the
regularization based on the acceleration error. In particular, for controlling
the end-effector to a fixed Cartesian point the forgetting factor converges to
1 (no forgetting; zero regularization) and within a couple second (including
  approach slowdown) and we achieve accuracies of around $10^{-5}$ meter.

\subsection{Handling noisy acceleration measurements} \label{sec:NoisyAccelerationMeasurements}
Section \ref{sec:onlineLearning} described the tools we use from the machine
learning (especially stochastic gradient descent) literature to reduce
oscillations. But additionally, since handling noisy data is a fundamental
problem to machine learning in general, these same tools enable us to handle
noisy acceleration measurements. 

Firstly, the basic algorithms, themselves, are robust to noise. These
gradient-based algorithms are most commonly applied in stochastic contexts,
where it is assumed that gradient estimates are noisy. And
Section~\ref{sec:TelescopingFiniteDifferences} discusses how sequential 
finite-differenced accelerations actually telescope to an extent allowing
the noise to inherently cancel over time.

But secondly, momentum acts as a damper to the forces generated by the
objective. Its interpretation as an exponential smoother shows that white noise
in the estimates cancels over time through the momentum as well, averaging to a
more clear acceleration signal.

And finally, we get 1000 training points per second. Physically the robot
doesn't move very far in a half a second, and we can assume that the errors in
the dynamics model will be changing at a time scale of tenths of a second, .1,
.5, or even 1 second. That's anywhere between 100 and 1000 training examples
available to track how errors in the dynamics model change as the robot
executes its policy, which is plenty of data to average out noise and run a
sufficient number of gradient descent iterations.

\subsection{A note on step size gains and an analogy to PD gains}

The larger the step size, the quicker the adaptive control strategy adjusts to
errors between desired and actual accelerations. That means it will fight
physical perturbations of the system stronger with a larger step size. To
accurately track desired accelerations, we either need large step sizes for
fast adaptation, or a good underlying dynamics model. If we have a really good
dynamics model, we can get away with smaller step sizes. That means the better the
dynamics model, the easier physical interaction with the robot becomes. Bad
models require larger step sizes which manifests as a feeling of ``tightness''
in the robot's joints. We can always push the robot around, and it'll always
follow its underlying acceleration policy from wherever it ends up, but the
more we rely on the adaptive control techniques, the tighter the robot becomes
and the more force we need to apply to physically push it around.

This behavior parallels the trade-offs we see in the choice of PD gains for
trajectory following. Bad (or no) dynamics models require hefty PD gains, which
means it can be near impossible to perturb the robot off it's path. But the
better the dynamics model, the better the feedforward term is, and the looser
we can make the PD gains while still achieving good tracking. We're able to
push the robot around more easily (and it's safer). The difference is whether
or not we need that trajectory. The proposed direct adaptive control method
attempts to follow the desired acceleration policy well, which means when we
perturb it, it always continues from where it finds itself rather than
attempting in any sense to return to a predefined trajectory or integral curve.

\section{Experiments}
\label{sec:experiments}
We evaluate our proposed approach both in simulation and on real world
control problems. We start out with illustrating our approach on a
simple 2D simulation experiment. Then, we show initial results of the
online learning in a Baxter simulation experiment. Finally, we move to
a real robotic platform - a KUKA lightweight arm - to illustrate the
effectiveness of our approach on a real system.
\subsection{A simple simulated experiment}
\begin{figure*}[t]
\begin{center}
\includegraphics[height=.235\columnwidth]{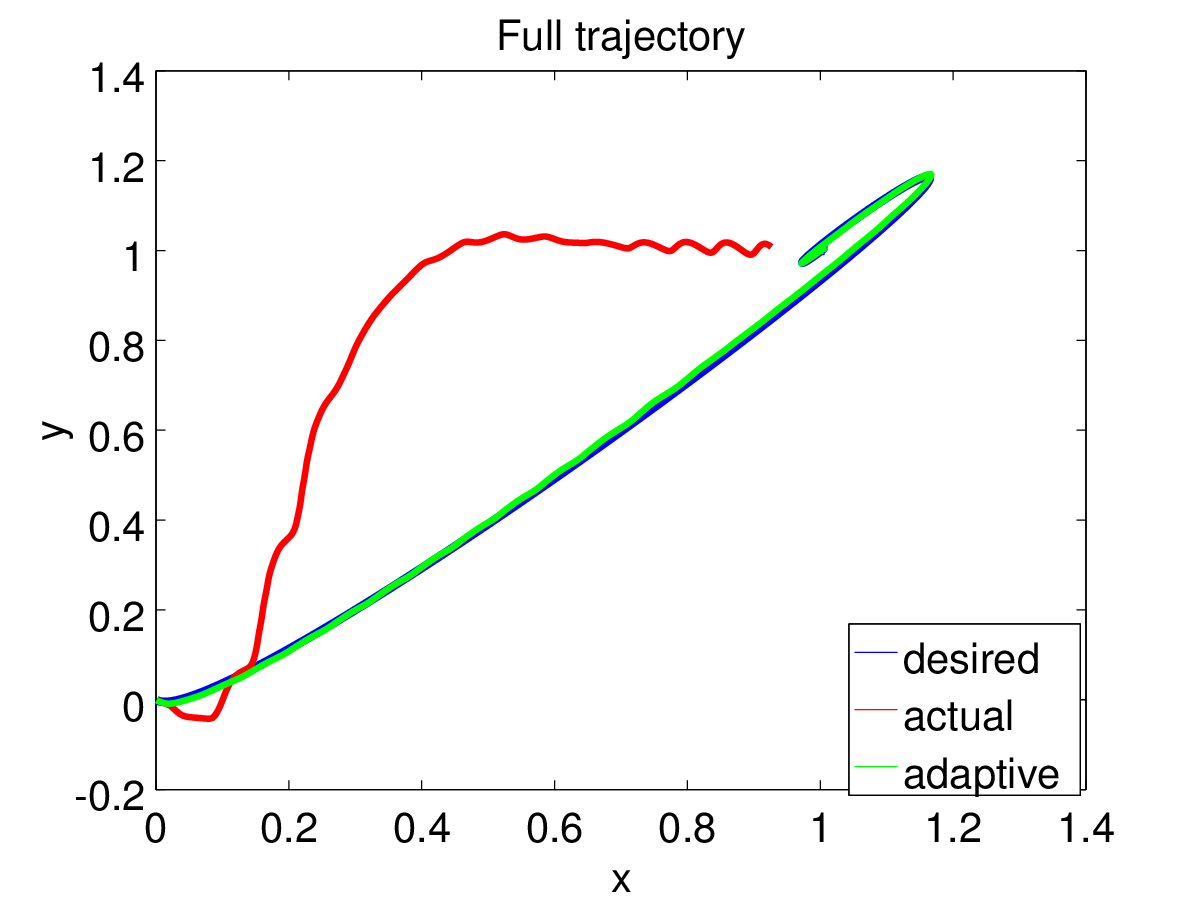}
\includegraphics[height=.235\columnwidth]{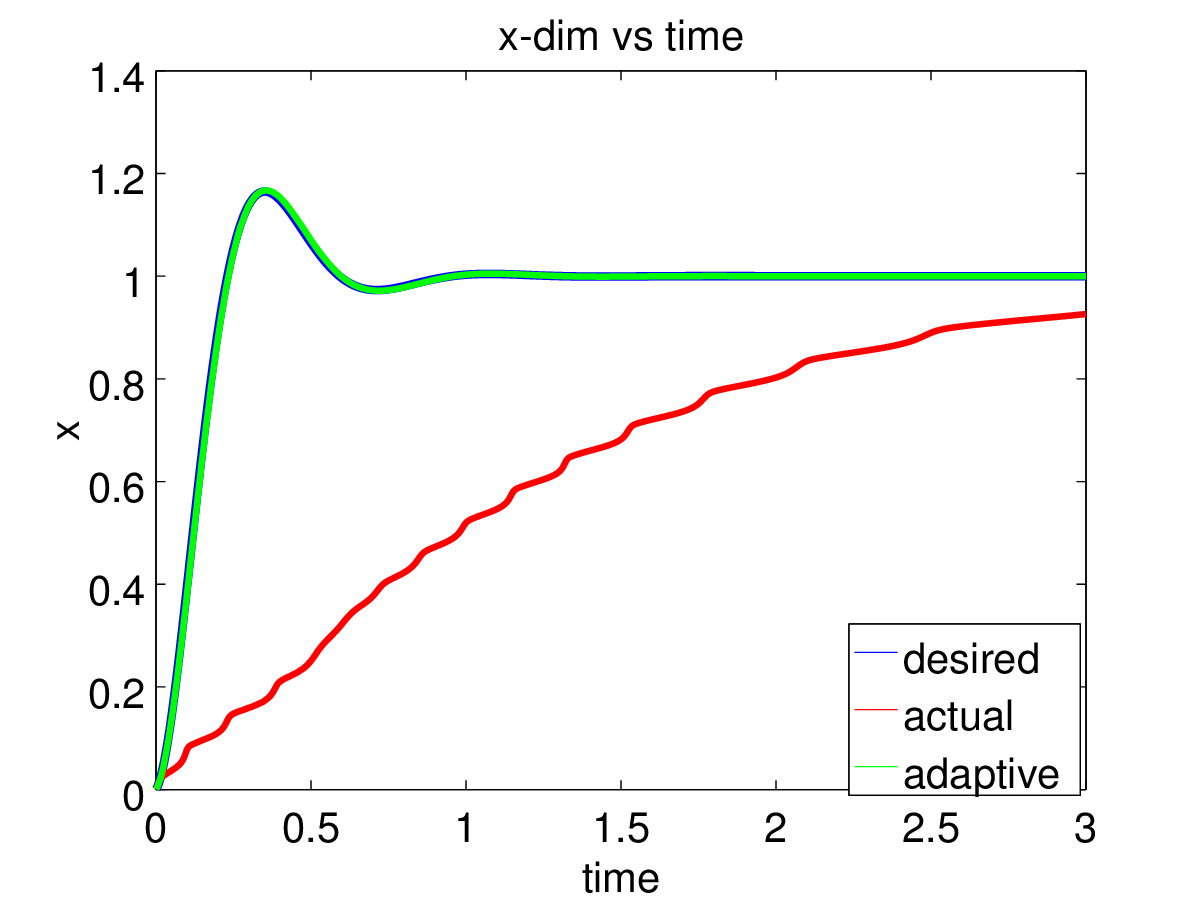}
\includegraphics[height=.235\columnwidth]{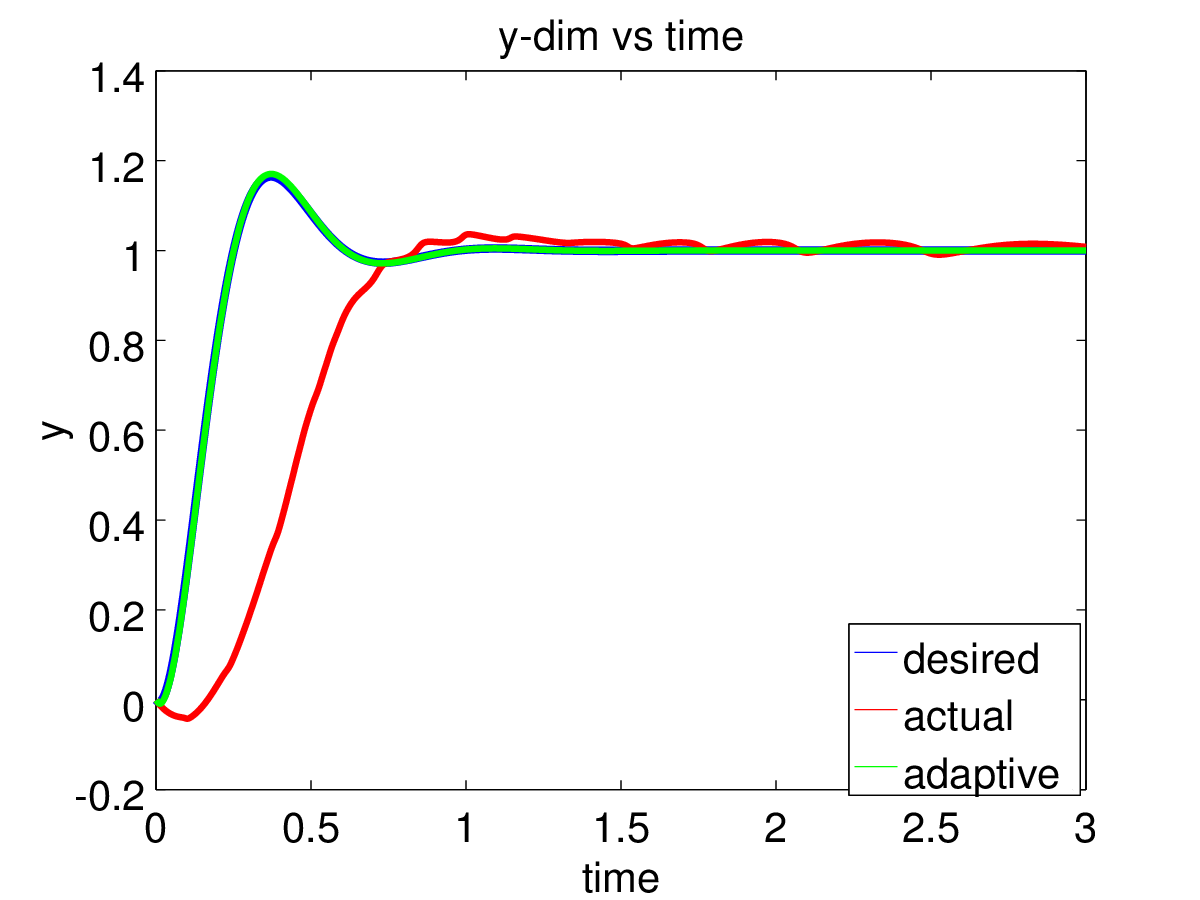}
\caption{\small 2D simulation of online adaptation to severe model inaccuracies and 
underlying friction (which varies sinusoidally with each dimension in this case). The time 
axes (lower axis of the second and third plots) are in units of seconds, and all spacial
axes are in units of meters. The controller updated at a rate of 1kHz.}
\label{fig:SinusoidalFrictionAdaptiveControl}
\end{center}
\end{figure*} 

This experiment shows a simple 2D example of the behavior of the adaptive
control system for a scenario where the dynamics model used by the robot
differs drastically from the true dynamics and where unmodeled nonlinear
frictions are significant.

The true mass matrix of the system is defined as $\M(\q) = 5 \big(\mv(\q) \mv(\q)' + .05 \I\big)$,
with \mbox{$\mv(\q) = (\sin(5q_1); \cos(2q_2))$}. The robot uses a diagonal constant
approximation of the form $\wideM = .5\I$, which assumes masses
that are an order of magnitude smaller. 
Additionally, the true system
experiences sinusoidal frictional forces of the form 
\begin{align}
  \mu(\q) = \left[
    \begin{array}{cc}
      100 \sin(50\ q_1) \\
      5 \sin(50\ q_2)
    \end{array}
  \right],
\end{align}
for which the robot has no knowledge of.

The system is using a simple PD controller (outputting accelerations) to move to
a desired fixed point to a target velocity of zero. 
\begin{align}
  \qdd = K (\q_d - \q) - D \qd,
\end{align}
where $K = 100.$ and $D = 10.$. The desired target is $\q_d = (1;1)$, and the robot 
starts from $\q_0 = (0;0)$ with a random initial velocity.

The acceleration controller itself is tuned incorrectly and therefore
overshoots. However, that's the desired behavior we want to track (shown in
blue).  If we simply pipe the desired accelerations through the very
approximate inverse dynamics model, the behavior we get is abysmal (shown in
red). On the one hand the dynamics model is a severe approximation, and on the
other hand, we have no advanced knowledge of the strange frictional pattern, so
the robot drifts upward and then oscillates during the final approach. But when
we turn on adaptation (shown in green), the system is able to compensate for
all of that and we get very good tracking behavior.

This implementation didn't simulate noise. But the real-world implementation
discussed in the next section shows the efficacy of the above described
learning tricks under noisy acceleration estimates sent back by the physical
robot.

\subsection{Experiments on a simulated Baxter platform}

\begin{figure*}[t]
 \centering
\includegraphics[trim=0mm -20mm -20mm 0mm, clip, width=.32\columnwidth]{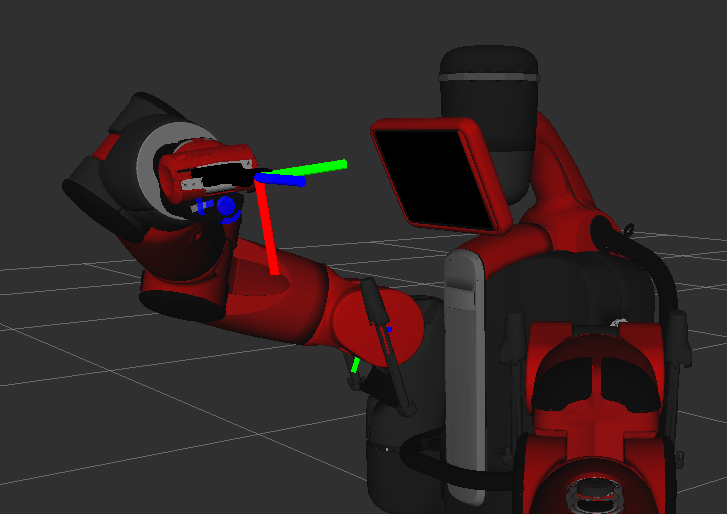}%
\includegraphics[trim=0mm 0mm 5mm 0mm, clip,width=.34\columnwidth]{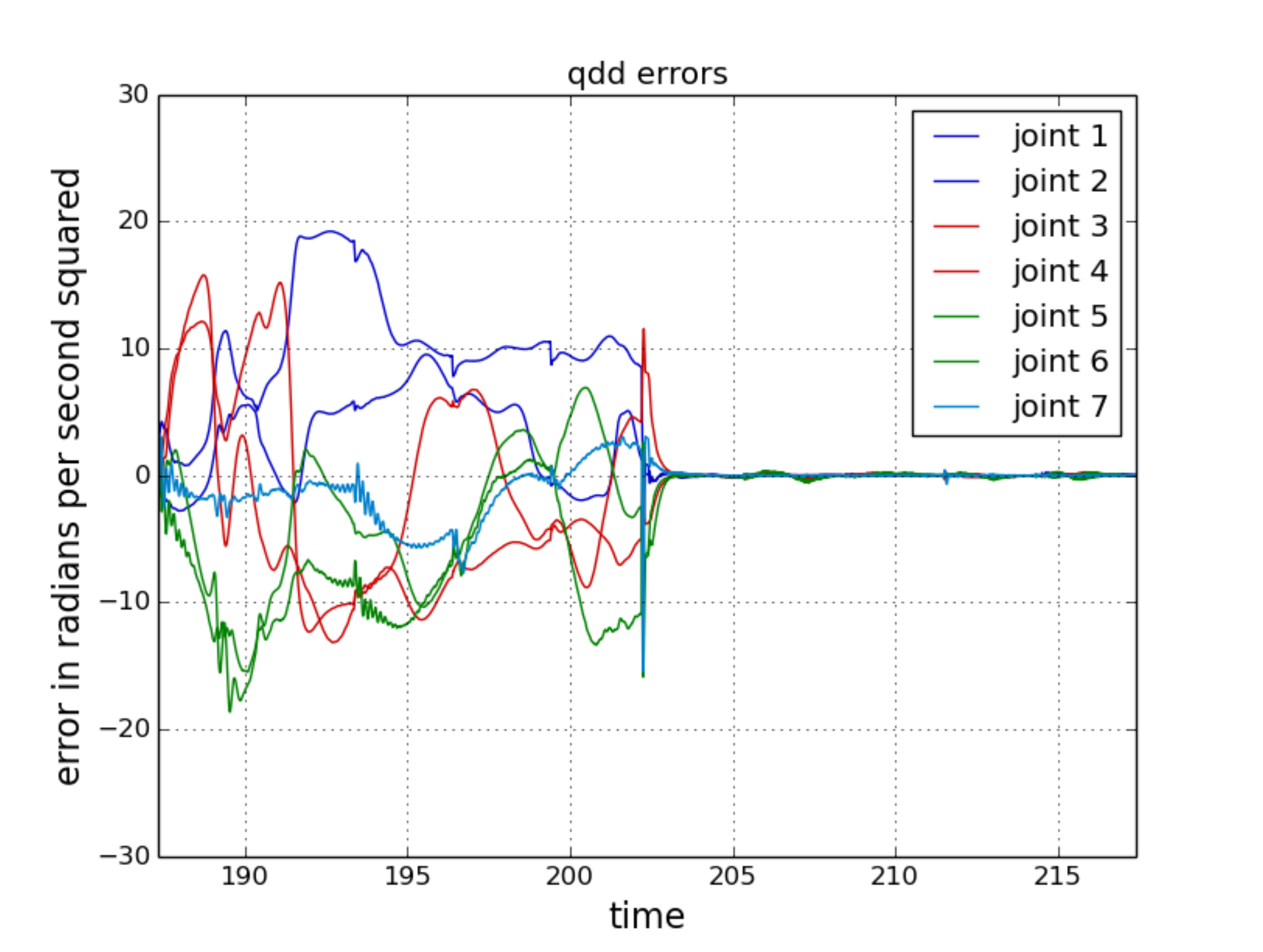}%
\includegraphics[trim=5mm 0mm 5mm 0mm, clip,width=.34\columnwidth]{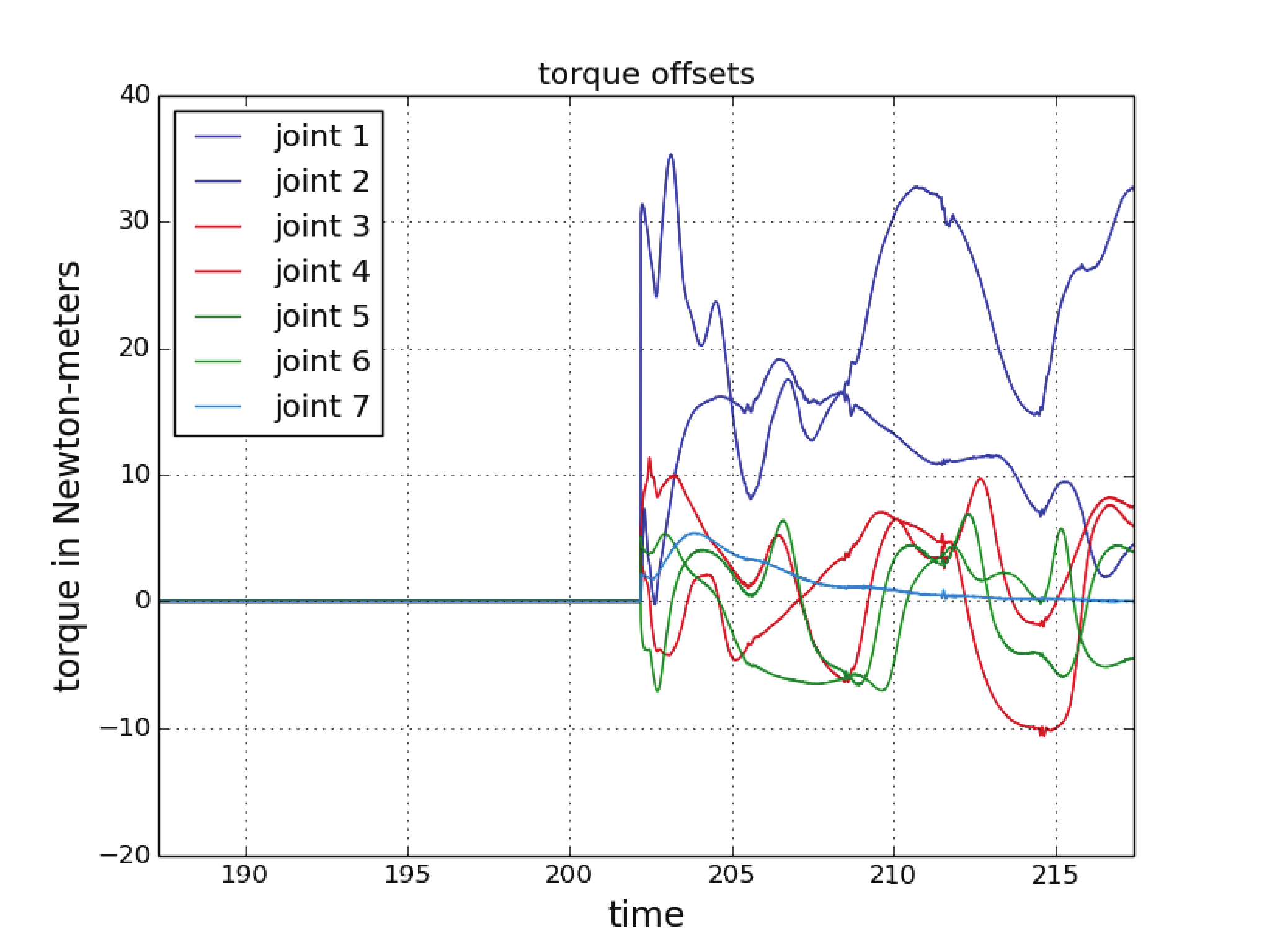}
\caption{\small Left: The Baxter robot. The middle and right plots depict the
acceleration errors and learned torque offsets of a 30 second run under severe
nonlinear torque biasing wherein online learning was switched on half way
through.}
\label{fig:Baxter}
\end{figure*} 

We ran a simulation experiment on the Baxter platform to analyze the
performance
of the online torque offset learner under severe modeling mismatches. The
controller
used inverse dynamics, but the simulator added substantial positionally 
dependent nonlinear friction and torque biases of the form
\begin{align*}
  \tau_{\mathrm{friction}} = -7 \sin^2(5\q) \Big(2\sigma(\qd) - 1\Big)
  \ \ \ \ \ \ 
  \tau_{\mathrm{bias}} = -5 \sin(5\q),
\end{align*}
where $\sigma$ is the typical 0-1 sigmoidal squashing function. Additionally,
the robot's internal model did not account for joint damping
coefficients,
while the simulator did.
The controller ran with a control cycle of $.005s$, and the simulator 
ran with an update cycle time of $.001s$.
We ran the controller through a 30 second sequence of 10 chained 3 second
Linear
Quadratic Regulators (LQRs), and turned on the online learning only half way
through.
Figure~\ref{fig:Baxter} shows severe acceleration errors resulting from the
model
mismatch during the first half, but half way through, the online learning turns
on and is able to track the torque offsets quite well, effectively zeroing the 
acceleration errors. The simulator added noise of the form 
$.001 \mathrm{uniform(-1,1)}$ to both the positions and offsets when reporting 
them to the controller, so the raw acceleration measurements were quite noisy.
The acceleration errors in the middle plot of the figure were, therefore,
exponentially
smoothed with a smoothing factor of $.975$. The online learner used momentum
updates, which is effectively a form of smoothing, so the final plot shows the
raw unsmoothed learned torque offsets used by the controller.

Note, that we have performed similar experiments both in our 2D simulated
inverse dynamics setup and on our KUKA lightweight arm, which all conclusively
confirm the effectiveness of the direct online learning approach. 

\subsection{Real-world experiments on a KUKA lightweight arm}
\label{sec:RealWorldExperiments}
\begin{figure*}
\begin{center}
  \includegraphics[width=0.45\linewidth]{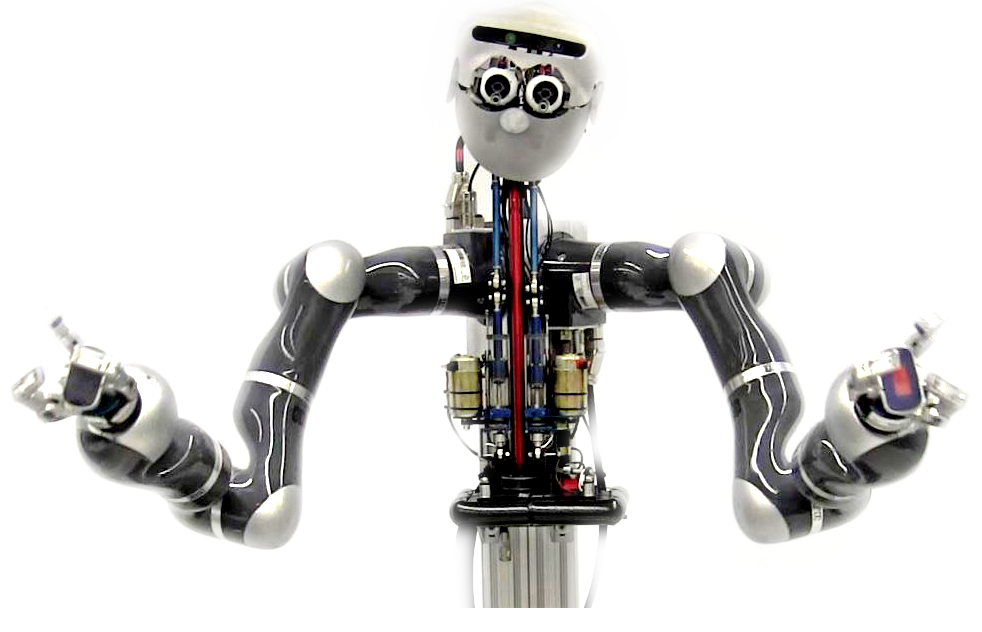}%
  \hfill
  \includegraphics[width=0.5\linewidth]{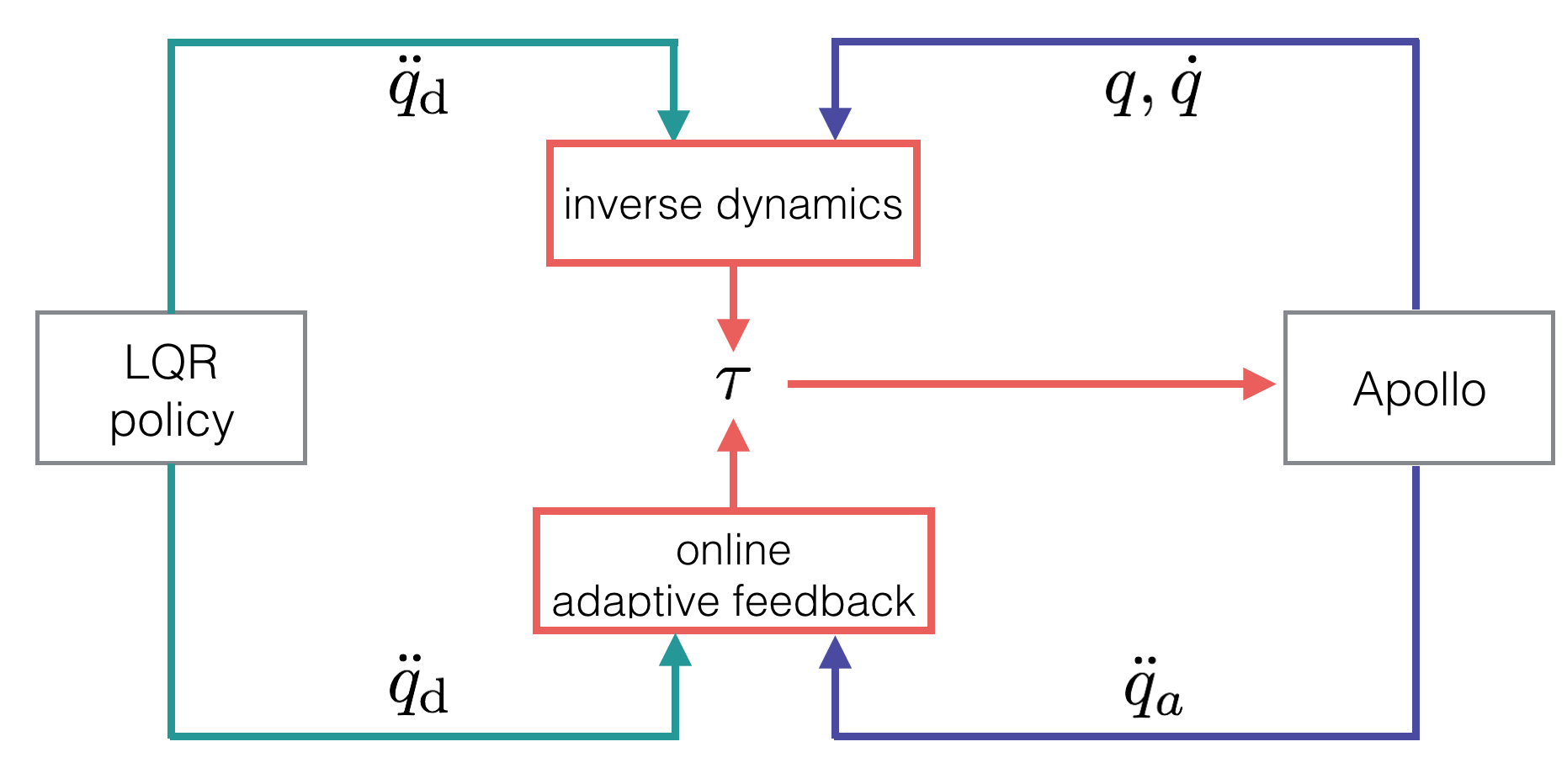}
\caption{\small (left) Apollo: our experimental platform with two KUKA
  lightweight arms. (right) The controller used for our experiments at
  a rate of 1kHz.}
\label{fig:experimental_setup}
\end{center}
\end{figure*}

We have a full implementation of this algorithm working for the Apollo
platform (see Figure~\ref{fig:experimental_setup} (left)) for both simple
Cartesian space controllers and for a full continuous optimization
(MPC-type) motion generation system.
The low level controller consists of an inverse dynamics controller,
that evaluates the (approximate) rigid body dynamics model for desired
accelerations. This inverse dynamics torque command is combined with
the offset estimated by the adaptive controller, see
Figure~\ref{fig:experimental_setup}. In all our experiments presented
here, we model the offset as a simple constant
$\f_\text{offset}(\q, \qd, \qdd_d, \w) = \w$. The KUKA lightweight arm
has $7$ degrees of freedom, thus the offset torque is a
$7$-dimensional vector.

To be robust against
noise, the adaptive controller transforms the gradient
$\qdd_d - \qdd_a$ as discussed in
Section~\ref{sec:algorithmDerivation}. The key parameters that are
involved for the offset update are
\begin{itemize}
\item the learning rate $\eta$ (Equation~\ref{eq:gradient_descent_step}),
\item the variance gain $\alpha$ (Equation~\ref{eq:variance_scaling}) and
\item the smoothing parameter $\gamma$ (Equation~\ref{eq:exponential_smoothing})
\end{itemize}
These parameters are shared across all joints. We start out with
analyzing the sensitivity of our proposed approach on the real Apollo
platform.
\subsubsection{Parameter Sensitivity Analysis}
\label{sec:param_sensitivity_analysis}
For our parameter sensitivity analysis we attempt to execute a
sequence of two pre-planned LQRs. This sequence has been executed three
times for each parameter combination. The considered values are:
\begin{itemize}
\item $\eta = 0.01, 0.02, \dots, 0.1$
\item $\alpha = 0.0, 0.1, \dots, 0.5$
\item $\gamma = 0.9, 0.95$
\end{itemize}
We have tested all combinations of these parameter settings and report
the average acceleration error per joint for both LQRs (averaged over
the three trials). If one LQR execution resulted in a unsuccessful
execution (within any of the 3 trials), that parameter setting was
classified as unsuccessful.
We evaluate the performance and sensitivity of our approach with the
help of three quantities:
\paragraph{The mean absolute acceleration error} between the desired
and actual accelerations, $\sum_t |\qdd^t_d - \qdd^t_a| /T$, where
T is the length of the movement. We plot this error as a function of
the learning rate $\eta$ and the variance gain $\gamma$. This plot is
color coded dark green to dark red. dark green indicates a very low
error, dark red higher errors. 
\paragraph{The mean acceleration error}
$\sum_t (\qdd^t_d - \qdd^t_a) /T$. A value of $0$ here would mean
on average the actual accelerations are neither under nor over the
desired acceleration. This plot is color coded from red to blue. Red
means, on average, we measure actual accelerations over the desired,
blue means we measure actual accelerations below the desireds. Darker
colors indicate a higher bias. Yellow colors represent a bias close $0$.
\paragraph{The mean absolute magnitude} of the adaptive torque command computed $\sum_t \flearn^t / T$. The magnitude of torque offset the adaptive controller adds to the torque command. Darker color indicates larger magnitudes.\\

\begin{figure*}
\begin{center}
\includegraphics[height=0.2\textheight]{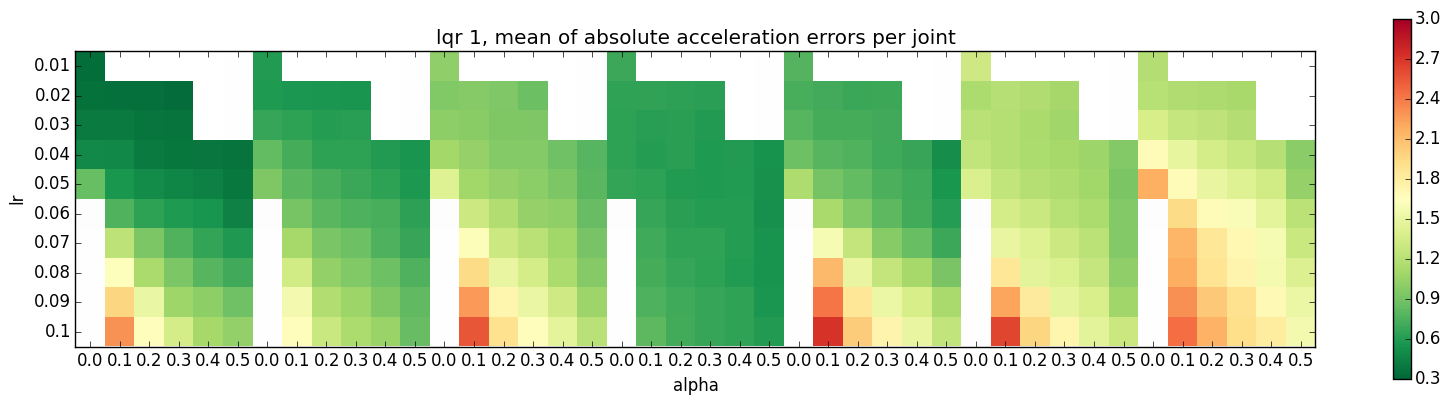}\\
\includegraphics[height=0.2\textheight]{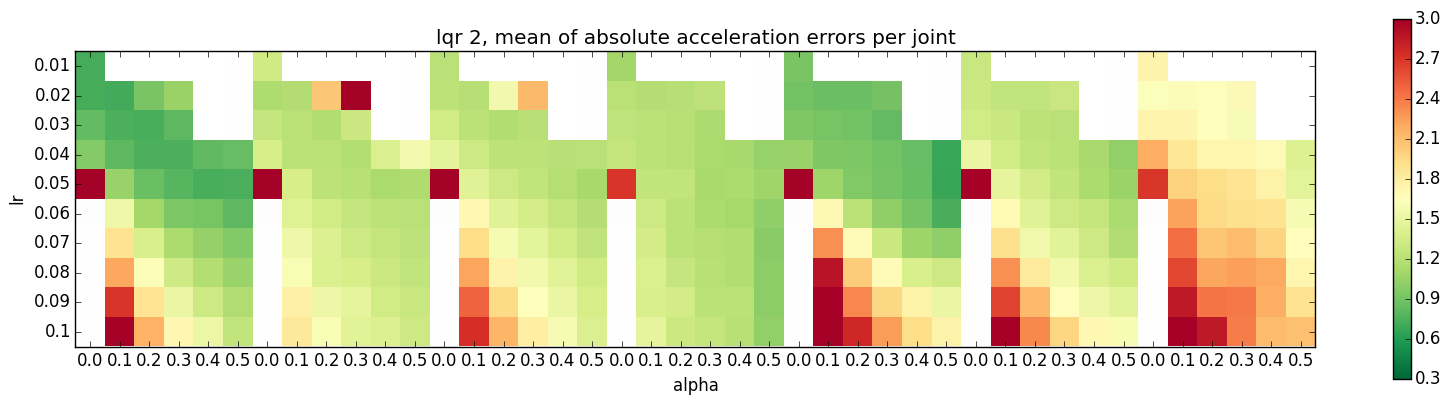}\\
\includegraphics[height=0.2\textheight]{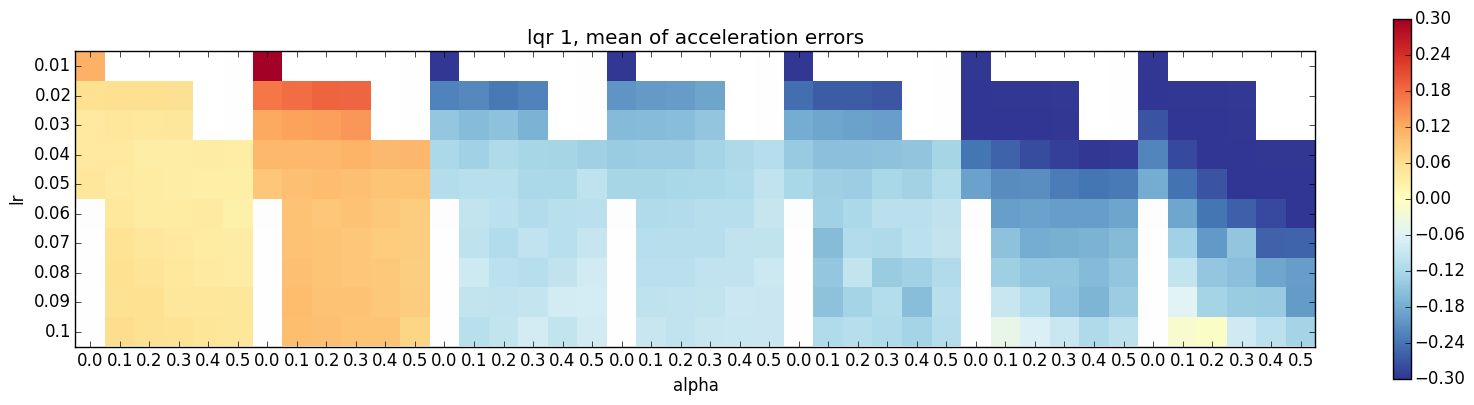}\\
\includegraphics[height=0.2\textheight]{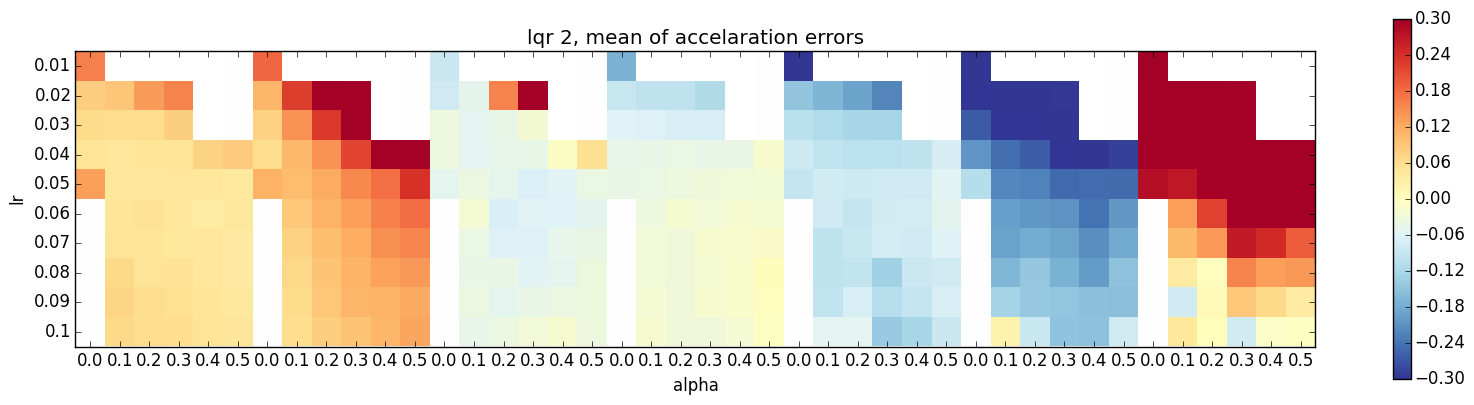}\\
\caption{\small The average acceleration error and the error bias for both LQRs with $\gamma = 0.9$. Results are displayed per joint, from left to right joint 1 to 7. From top to bottom: the average absolute acceleration for LQR \#1 and \#2, the mean of the acceleration error bias for both LQRs.}
\label{fig:param_sensitivity1}
\end{center}
\end{figure*}

\begin{figure*}
\begin{center}
\includegraphics[height=0.2\textheight]{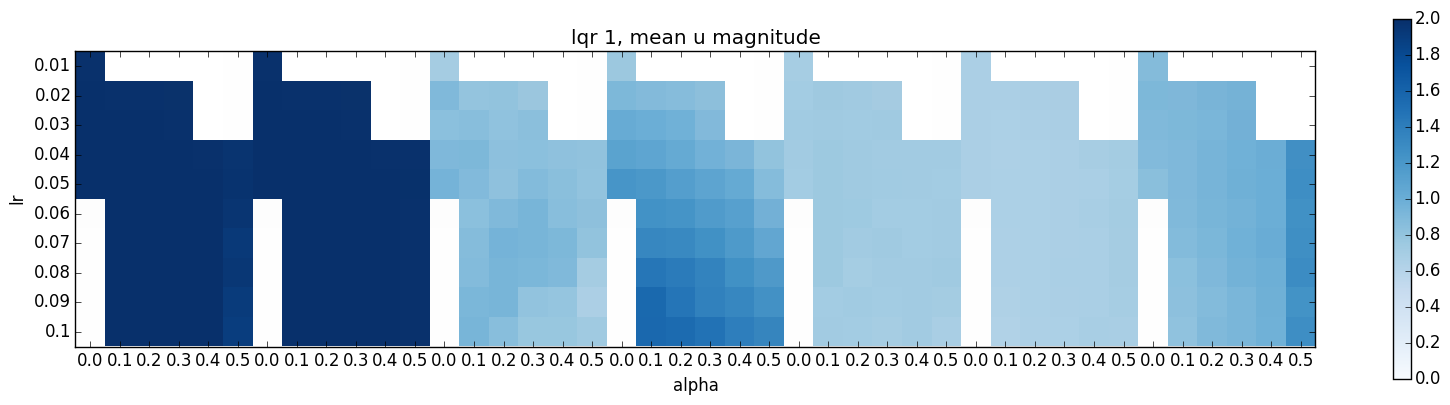}\\
\includegraphics[height=0.2\textheight]{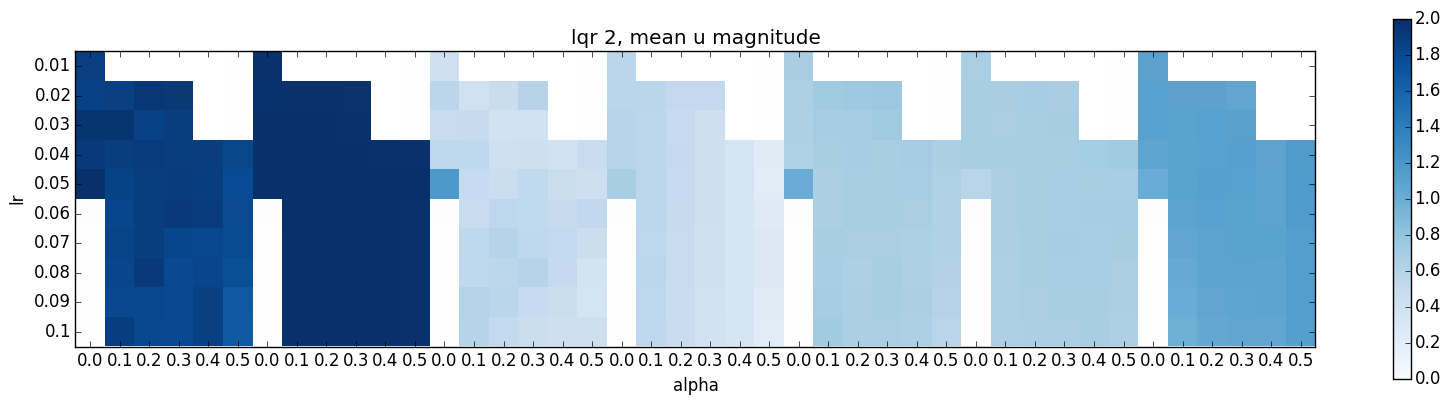}%
\caption{\small Torque magnitudes estimated by the adaptive
  controller $\gamma = 0.9$. (top) LQR \#1, (bottom) LQR \#2. Results are shown  per joint, from left to right joint 1 to 7.}
\label{fig:adaptive_torque_offsets1}
\end{center}
\end{figure*}

In Figure~\ref{fig:param_sensitivity1} we plot the average absolute
acceleration error and the bias of the acceleration error, for all
combinations of $\eta$ and $\alpha$ while keeping $\gamma=0.9$ fixed.
The first result to notice is that we were able
to run the controller for a large portion of parameter combinations.
While the error varies across the parameter settings, we can deduce
that even without an additional error feedback term, we don't have to
perform excessive tuning of the parameters to obtain an (empirically)
well behaved controller. Note, that in all failure cases (indicated by
white squares in the plots) it was typically during the execution of
the 2nd LQR that the movement was deemed unsafe.
The second result is that there seems to exist a relative large band
across the 2 parameter settings that seems to achieve low acceleration
errors (top two plots of Figure~\ref{fig:param_sensitivity1}). This
promises that tuning the adaptive controller is not too difficult.
When looking at the mean acceleration error bias we notice that we
seem to typically overestimate the required torques for joints $1$ and
$2$ (the shoulder joints) and underestimate the required torque for
joints $3-6$.
This is best understood, by keeping in mind that we are fixing the
approximate RBD model.
Depending on the movement this approximate model may over or
underestimate the required torque to achieve desired accelerations.
If the adaptive controller can fix the model - the bias should become
smaller (which is represented by lighter colors).
For low learning rates, we tend to observe larger bias values,
This indicates that we may not be adapting to the error
fast enough, and we consistently over/underestimate the required torque
offsets, as can be seen in Figure~\ref{fig:param_sensitivity1}, for joints
2-4 for instance.
However, simply increasing the learning rate is not necessarily the
right thing to do. Notice, although for joint 1 we observe a small
bias of the acceleration error for higher learning rates, the average
absolute error increases as we increase the learning rate. This tells
us that there may be oscillations, that on average have a small bias.
This can mean that we are too aggressive in trying to fix the model,
incurring larger acceleration errors, that we try to account for at the
next time step. The severity this effect seems to also be a function
of the variance gain. With smaller values this effect seems to
pronounced (for instance, see joint 1, LQR \# 2 the bottom left corner
for both error measures).
Finally, to make sure that our adaptive torque offset estimates are
sensible, we plot the offset torque magnitude per joint in
Figure~\ref{fig:adaptive_torque_offsets1}. We observe, that for the
should joints we estimate higher torque offsets than for the the elbow and wrist joints. In general the offset torques range between $0 Nm$ to $2 Nm$, which is a reasonable range.

We have repeated the same set of experiments with a smoothing parameter of $\gamma=0.95$. The figures for this set of experiments can be found in the appendix \ref{sec:more_plots}. In general, the relationship between the variance gain and learning rate is similar to the above presented results. However, we notice that the the effect of higher learning rates is more extreme for some joints. For instance, for joint 5, the acceleration error seems to increase faster (as the learning rate is increased), indicating that oscillations are a function of both the learning rate and smoothing parameter. Intuitively, this makes sense, as the smoothing slows down the response to the errors as well, thus a high learning rate with a very little smoothing seem to not be an optimal pairing.
All in all we can conclude that the adaptive controller can
effectively fix errors in the dynamics model on the fly. Tuning the
parameters, while unavoidable, is not too difficult.

\subsubsection{Video demonstrations}
The following three videos show examples of the
dynamics adaptation algorithm running on the Apollo manipulation
platform. In the first two videos, Apollo is directly executing an
acceleration policy designed to generate Cartesian motion moving his
finger to a fixed point in space. In the final video, Apollo is
executing optimized grasping behaviors.
\begin{enumerate}
\item {\bf Robustness to perturbation:} \url{https://youtu.be/clldz75ToVI} \\
In this video, Apollo is repeatedly perturbed away from the fixed point and 
allowed to return under the control of the acceleration policy. The behavior 
is shaped by the desired accelerations, which Apollo is able to accurately reproduce
by running the online learning adaptive control method outlined above 
to track how the dynamics model error shifts throughout the execution.
\item {\bf Bounce tests:} \url{https://youtu.be/m0i5oHQeqA8} \\
In this video, Apollo is put through a series of more aggressive bounce tests 
in an attempt to incite oscillation modes. The robustness measures outlined
above successfully combat the maneuvers and Apollo's behavior remains
natural throughout the attempt.
\item {\bf Grasp tasks:} \url{https://youtu.be/LQABeK2IO80} \\
In this video, Apollo is executing grasp task motions optimized on
on the fly. Each motion is sent down to control as a sequence of affine 
acceleration policies which are directly executed using the online learning 
adaptive control methodology described in this document.
No vision is used; the sequence of object locations is planned out in advance.
The system, however, uses force control in the grasps (reading from the fingertip's 
strain gauges) to be robust to variations in size and specific positioning 
of the objects.
\end{enumerate}
The underlying adaptive control parameterization is the same in all
cases---we tune once and that enables the execution of any number of
acceleration policies.

\section{Theoretical connections and notes}
\label{sec:theoretical_connections}

This section presents some additional theoretical results to give the direct
loss adaptive control methods discussed in this paper context. We start by
using the techniques from above to derive a related loss function whose
constant step size online gradient descent is PID control. This analysis shows
that we can additionally reinterpret classical algorithms as forms of online
gradient descent on concrete loss functions, and that we can potentially
leverage these techniques to generalize PID control to be more adaptive by
utilizing modern tools from the machine learning literature.

We also present a result that addresses some of the questions around why we can
use finite-differenced accelerations in practice, despite their noise. We show
that gradient descent on the acceleration loss presented above (the simplest
constant step size variant) can be viewed as a form of virtual velocity
control---by expanding the terms of the algorithm and rearranging them, the
finite-differenced accelerations, to a large extent, telescope into a nice
compact virtual velocity estimate over time. Despite the high variance of the
finite-differenced acceleration estimates themselves, the final expression
collapses into a form dependent on just velocity measurements (which are either
directly measurable or at most first-order differenced), showing that the
variance of the control law itself is actually much lower than would be
expected from a simple additive noise argument under finite-differenced
accelerations.

\subsection{A loss function for PID control} \label{sec:LossFunctionPID}

In much of this paper, we have assumed that the robot is following a feedback
acceleration policy producing a steady stream of (only) desired accelerations
as a function of state $\qdd_d = \f(\q, \qd)$.  In this section, we show how we
can leverage similar ideas to write out related objective terms measuring
errors in position and velocity as well. We'll see that one particular variant
of the resulting online learning algorithm is the traditional PID control used
frequently in many real-world systems.

We've already seen in Section~\ref{sec:algorithmDerivation} 
an online learning algorithm for 
directly minimizing acceleration errors given desired acceleration signals
$\qdd_d$. This section derives similar objective terms that measure the error
relative to desired position $\q_d$ and velocity $\qd_d$ signals as well.

\subsubsection{Position error term}

Executing an acceleration $\qdd_a^{t}(\w)$ for $\Delta t$ seconds from $\q$
(moving at velocity $\qd$)
results in a new position given by
\begin{align}
  \q_a^{t+1}(\w) = \q^{t} + \Delta t\ \qd^{t} + \frac{1}{2} \Delta t^2 \qdd_a^{t}(\w).
\end{align}
If we desire to be at $\q_d^{t+1}$ \textit{after} executing an actual measured acceleration 
$\qdd_a^{t}(\w)$ from the current state, the resulting position error can, therefore, be expressed as
\begin{align} \label{eqn:positionLoss}
  l_\textrm{pos}^{t+1}(\w) = \frac{1}{\Delta t^2}\|\q_d^{t+1} - \q_a^{t+1}(\w)\|_{\M_t}^2,
\end{align}
where $\M_t$ is the true mass matrix which defines the real system dynamics 
at the time of executing the acceleration. The derivative is
\begin{align}
  \nabla_\w l_{\mathrm{pos}}^{t+1}(\w)
  &= -\frac{2}{\Delta t^2}\left[\frac{\partial\q_{a}^{t+1}(\w)}{\partial\w}\right]^T\M_t\Big(\q_d^{t+1} - \q_a^{t+1}(\w)\Big) \\
  &= -\J_{f_t}^T\Big(\q_d^{t+1} - \q_a^{t+1}(\w)\Big),
\end{align}
since $\frac{\partial\q_{a}^{t+1}(\w)}{\partial\w} = \frac{1}{2}\Delta t^2 \M_t^{-1}\J_{f_t}$.
This gradient is the position error pushed through the Jacobian.

\subsubsection{Velocity error term}

Similarly, 
executing acceleration $\qdd_a^t(\w)$ for $\Delta t$ seconds from velocity $\qd^t$
results in a new velocity given by
\begin{align}
  \qd_a^{t+1}(\w) = \qd^t + \Delta t\ \qdd_a^t(\w).
\end{align}
If we desire to be at velocity $\qd_d^{t+1}$ after executing an actual measured acceleration 
$\qdd_a^t(\w)$ from the current state, the resulting velocity error can, therefore, 
be expressed as
\begin{align} \label{eqn:velocityLoss}
  l_\textrm{vel}^{t+1}(\w) = \frac{1}{2\Delta t}\|\qd_d^{t+1} - \qd_a^{t+1}(\w)\|_{\M_t}^2,
\end{align}
where again $\M_t$ is the true mass matrix which defines the real system dynamics 
at the time of executing the acceleration. The derivative is
\begin{align}
  \nabla_\w l_{\mathrm{vel}}^{t+1}(\w)
  &= -\frac{1}{\Delta t}\left[\frac{\partial\q_{a}^{t+1}(\w)}{\partial\w}\right]^T\M_t\Big(\qd_d^{t+1} - \qd_a^{t+1}(\w)\Big) \\
    &= -\J_{f_t}^T\Big(\qd_d^{t+1} - \qd_a^{t+1}(\w)\Big),
\end{align}
since $\frac{\partial\q_{a}^{t+1}(\w)}{\partial\w} = \Delta t \M^{-1}_t\J_{f_t}$.
This gradient is the velocity error pushed through the Jacobian.

\subsubsection{The PID objective}

Combining these two terms with the acceleration error term derived in 
Section~\ref{sec:algorithmDerivation}, gives an objective of the form
\begin{align} \label{eqn:PIDObjective}
  l^t(\w)
    &= \alpha\ l_\mathrm{pos}^t(\w)
      + \beta\ l_\mathrm{vel}^t(\w) 
      + \gamma\ l_\mathrm{acc}^t(\w)  \\
    &= \frac{\alpha}{\Delta t^2}\|\q_d^t - \q_a^t(\w)\|_{\M_t}^2
      + \frac{\beta}{2\Delta t}\|\qd_d^t - \qd_a^t(\w)\|_{\M_t}^2
      + \frac{\gamma}{2}\|\qdd_d^t - \qdd_a^t(\w)\|_{\M_t}^2,
\end{align}
where $\alpha$, $\beta$, and $\gamma$ are scaling constants.
The gradient is given by
\begin{align}
  \nabla_{\w}l^t(\w_t) 
  = -\J^T_{f_{t-1}}\Big(\alpha\left(\q_d^t - \q_a^t\right)
      + \beta\left(\qd_d^t - \qd_a^1\right)\Big)
      + \gamma\J^T_{f}\left(\qdd_d^t - \qdd_a^t\right),
\end{align}
where $\J_{f_t}$ is the Jacobian of the offset function at $\w_t$.

A number of gradient-based algorithms can be derived from this gradient expression.
But in particular, consider
$\flearn(\w) = \w$ such that $\J_{f_t} = \I$ for all $t$. Then
summing these gradients under a constant step size of $\Delta t$, which implements
a vanilla constant step size gradient descent method, results in the following 
control law:
\begin{align}
  \w^{t} &= \torq^{t}
  = -\sum_{\kappa=0}^{t-1}\nabla_\w l^\kappa(\w^\kappa) \\
  &= 
  \alpha\sum_{\kappa=0}^{t-1}\left(\q^\kappa_d - \q^\kappa_a\right) \Delta t
    + \beta\sum_{\kappa=0}^{t-1}\left(\qd^\kappa_d - \qd^\kappa_a\right) \Delta t
    + \gamma\sum_{\kappa=0}^{t-1}\left(\qdd^\kappa_d - \qdd^\kappa_a\right) \Delta t \\
    &= \alpha\sum_{\kappa=0}^{t-1}\left(\q^\kappa_d - \q^\kappa_a\right) \Delta t
    + \beta\left(\q^t_d - \q^t_a\right)
    + \gamma\left(\qd^t_d - \qd^t_a\right).
\end{align}
These three terms are the integral, position error, and velocity error
terms, respectively, of a PID controller. This shows that we 
may view PID control as gradient descent on 
the objective given by \ref{eqn:PIDObjective}
under a constant step size. Beyond that, it additionally shows that there 
is also a much broader class of learning algorithms we can leverage to achieve
higher fidelity tracking as we've seen above in our experiments
on the acceleration error term alone. Studying this broader class of PID-related
algorithms is an interesting avenue for future work.

\subsection{Telescoping finite-differenced accelerations: a virtual velocity control interpretation}
\label{sec:TelescopingFiniteDifferences}

In this section, we show that the simplest version of our algorithm, with a bit a rearranging,
can be written as a form of virtual velocity control. The finite-differenced accelerations,
which are worrisome to many due to possible estimation noise, telescope in the sum across
time into a relatively simple form of velocity control expression wherein target 
velocities are integrated forward from desired accelerations (with a forgetting factor)
and control signals are derived as velocity differences. We present this result first in 
equation form and discuss its implications, and then derive it algebraically.

\subsubsection{Basic result} \label{sec:BasicResult}

The simplest version gradient descent algorithm given in 
Equation~\ref{eq:gradient_descent_step}, using a constant step size, takes the form:
\begin{align} \label{eqn:OnlineLearningRule}
  \torq^{t+1} = 
    \alpha (\qdd_d^t - \qdd_a^t) 
    + (1-\alpha\wt{\lambda})\torq^t,
\end{align}
where $\alpha$ is the constant step size and $\wt{\lambda}$ is the regularization constant.
This algorithm is gradient descent using gradient 
$-(\qdd_d^t - \qdd_a^t) + \wt{\lambda} \torq^t$. For the analysis below,
it's
convenient to define $\alpha\wt{\lambda} = 1-\lambda$ so that the result
is more comparable to the next algorithm, which is why we use a slightly different notation
here than in Equation~\ref{eq:gradient_descent_step}.

We will show below that if we expand this recursive relationship and rearrange some 
terms, we can write this (constant step size) control law as
the form
\begin{align} \label{eqn:ClassicalRule}
  \torq^{t+1} 
  = \alpha \Big[
    \Big(
    \underbrace{\big(\lambda \mv^t + (1 - \lambda)\qd^t\big) + \Delta t \qdd_d^t}_{\mv^{t+1}}
    \Big)
    - \qd^t
  \Big].
\end{align}
Here $\mv^t$ is a virtual target velocity vector that's integrated forward
using the desired accelerations. At the beginning of each control cycle the
virtual velocity vector is blended slightly with the current state via a
forgetting factor and then integrated forward by one step using the desired
acceleration $\qdd_d^t$.

In all cases, $\qdd_d^t = \f(\q^t, \qd^t)$ is the desired acceleration
evaluated at the \textit{current} measured state.

\subsubsection{Discussion} \label{sec:EquivalenceDiscussion}

In general, leveraging finite-differenced accelerations is worrisome due to the
noise in the estimates. We've shown experimentally, however, that it can work
in practice for this particular algorithm, enabling accurate tracking of raw
acceleration policies. The result in
Section~\ref{sec:BasicResult} provides some insight into why.

At a high level, since we're summing across finite-differenced acceleration
errors over time, we're effectively integrating. The sum of true accelerations
becomes the true velocity and the sum of desired accelerations becomes a
virtual desired velocity (there are some details regarding the forgetting
factor (see the next section), but this is the basic mechanism at work). This
algorithm is, therefore, a form of velocity control where desired velocities
are generated from the underlying acceleration policy. In practice, we use more
sophisticated variants of gradient-based online learning that leverage varying
step sizes, momentum, or other tricks that offer increased adaptation in real
time. The basic structure of these algorithms is the same, and the overall
numerical process retains the same sort of telescoping form making it suitable
for practical execution.

\subsubsection{Equivalence} 

This section derives the equivalence result discussed in
Section~\ref{sec:BasicResult}. We can show the equivalence by expanding the
rules and writing them both as a difference between a virtual desired velocity
and the current velocity.  We start with the online learning variant.

\hfill \\
\noindent{\bf Direct loss adaptive control algorithm expansion}
\hfill \\

Using the notation
$\alpha\wt{\lambda} = 1-\lambda$,
we can 
rewrite the online learning rule in Equation~\ref{eqn:OnlineLearningRule}
as
\begin{align}
  \torq^{t+1}
  &= 
    \alpha (\qdd_d^t - \qdd_a^t) 
    + \lambda\torq^t \\
  &=
    \alpha\Big(
      \big( \qdd_d^t - \qdd_a^t \big)
      + \lambda\big( \qdd_d^{t-1} - \qdd_a^{t-1} \big)
      + \cdots
    \Big).
\end{align}
We see here that $\alpha$ is just a gain, so we can drop it without 
loss of generality. Now we expand the acceleration estimate 
in terms of its finite-differencing expression 
$\qdd^t = \frac{1}{\Delta t}\big(\qd^t - \qd^{t-1}\big)$:
\begin{align}
  \torq^{t+1} 
  &\propto
      \Big( \qdd_d^t -  \frac{1}{\Delta t}\big(\qd^t - \qd^{t-1}\big)\Big)
      + \lambda\Big( \qdd_d^{t-1} -  \frac{1}{\Delta t}\big(\qd^{t-1} - \qd^{t-2}\big)\Big)
      + \cdots
\end{align}
The key observation is that we almost have a 
telescoping sum in these finite-differencing terms. 
If $\lambda$ were $1$, the sum would telescope and result
in just $-\frac{1}{\Delta t}\qd^t$. Instead, we get some (significant) residual from those terms
and it's insightful to group them together:
\begin{align}
  \torq^{t+1} 
  \ \ \propto\ \ 
  &\qdd_d^t + \lambda \qdd_d^{t-1} + \lambda^2 \qdd_d^{t-2} + \cdots \\
  & - \frac{1}{\Delta t}\qd^t \\
  & +
  \underbrace{
    \left(
      \frac{1}{\Delta t} \qd^{t-1} - \frac{\lambda}{\Delta t} \qd^{t-1}
    \right)
    + \left(
      \frac{\lambda}{\Delta t} \qd^{t-2} - \frac{\lambda^2}{\Delta t} \qd^{t-2}
    \right)
    +\cdots.
  }_{\frac{1 - \lambda}{\Delta t}\Big(\qd^{t-1} + \lambda \qd^{t-2} + \cdots\Big)}
\end{align}
Collapsing these terms into summation expressions and pulling out the $1/\Delta t$
as a constant factor (and absorbing it into the proportionality), we get
\begin{align}
  \torq^{t+1} 
  \ \ \propto\ \ 
  \sum_{\kappa=0}^t \lambda^\kappa \Delta t\qdd_d^{t-\kappa}
  - \qd^t
  + (1-\lambda) \sum_{\kappa=0}^{t-1} \lambda^\kappa \qd^{t-\kappa - 1}.
\end{align}
Note that $\sum_{\kappa=0}^{\infty}\lambda^\kappa = \frac{1}{1 - \lambda}$.
Thus, as $t$ gets larger, $1-\lambda$ increasingly approximates
a normalization factor on weights $\lambda^\kappa$. So defining
$w^\kappa = (1 - \lambda)\lambda^\kappa$, it's increasingly accurate
to say $\sum_{\kappa=0}^t = w^\kappa \approx 1$, i.e. using the 
weights in a sum forms a exponential weighted average. Using this 
notation, and rearranging the terms slightly, we get
\begin{align}
  \torq^{t+1} 
  \ \ \propto\ \ 
  \underbrace{
  \Big(
  \sum_{\kappa=0}^{t-1} w^\kappa \qd^{t-\kappa - 1}
    + \sum_{\kappa=0}^t \lambda^\kappa \Delta t\qdd_d^{t-\kappa}
  }_{\wt{\mv}^{t+1}}
  \Big)
  - \qd^t.
\end{align}
We explicitly attribute the first two terms to the virtual velocity
$\wt{\mv}^{t+1}$. The intuition is that the first term 
$\sum_\kappa w^\kappa \qd^{t - \kappa - 1}$ is just a smoothed estimate of the velocity
found by calculating a exponentially decaying weighted average over the measured
velocities. The second term $\sum_\kappa \lambda^\kappa \Delta t \qdd_d^{t-\kappa}$,
on the other hand, is a vector defining (approximately) what velocity we 
should have gotten if we actually accelerated from $0$ as defined 
by the sequence of desired accelerations $\qdd_d^{t-\kappa}$.
Each $\Delta t \qdd_d^{t-\kappa}$ is what we'd add to the velocity in an integration 
step. If $\lambda=1$, this expression would exactly be the integration 
expression. But since $\lambda<1$, we actually have it fully using 
the first term $\Delta t \qdd_d^t$ accounting for the latest acceleration,
but increasingly forgetting 
past accelerations over time. That first sum expresses the velocity 
we should see starting from zero, so adding it to the weighted
average velocity defines where we actually end up starting from 
the (smoothed) measured velocity. That smoothed measured velocity
is constantly updating to the robot's current velocity, 
and the acceleration integration term is forgetting old
accelerations at the same rate (with exponential decay),
so the combined process stays up-to-date at complimentary rates.

This analysis shows that we can view the online gradient descent 
algorithm as a form of virtual velocity feedback control with a 
virtual velocity vector of the form:
\begin{align} \label{eqn:OnlineLearningVirtualVelocity}
  \wt{\mv}^{t+1} = 
  \sum_{\kappa=0}^{t-1} w^\kappa \qd^{t-\kappa - 1}
  + \sum_{\kappa=0}^t \lambda^\kappa \Delta t\qdd_d^{t-\kappa}
\end{align}
The next subsection shows that this is essentially the same expression as is
used by the classical algorithm.

\hfill \\
\noindent{\bf Virtual velocity feedback expansion}
\hfill \\

Now lets expand the virtual velocity expression in the classical update
rule of Equation~\ref{eqn:ClassicalRule}:
\begin{align}
  \mv^{t+1} 
  & = \Big[\lambda \mv^t + (1 - \lambda)\qd^t\Big] + \Delta t \qdd_d^t \\
  &= 
    \Big[
      \lambda\Big(
        \lambda \mv^{t-1} + (1-\lambda) \qd^{t-1} + \Delta t \qdd_d^{t-1}
      \Big)
      + (1-\lambda) \qd^t
    \Big]
    + \Delta t \qdd_d^t \\
  &=
    \lambda^2 \mv^{t-1}
    + \lambda(1 - \lambda) \qd^{t-1}
    + (1 - \lambda) \qd^t
    + \lambda \Delta t \qdd_d^{t-1}
    + \Delta t \qdd_d^t \\
  &=
    (1 - \lambda) \sum_{\kappa=0}^t \lambda^\kappa \qd^{t - \kappa}
    + \sum_{\kappa = 0}^t \lambda^\kappa \Delta t \qdd_d^{t-\kappa}.
\end{align}

Again, we can write $w^\kappa = (1-\lambda)\lambda^\kappa$ to get
\begin{align} \label{eqn:ClassicalVirtualVelocity}
  \mv^{t+1} = 
    \sum_{\kappa=0}^t w^\kappa \qd^{t - \kappa}
    + \sum_{\kappa = 0}^t \lambda^\kappa \Delta t \qdd_d^{t-\kappa}.
\end{align}
Comparing this expression to that for the virtual velocity
of the online learning update rule in 
Equation~\ref{eqn:OnlineLearningVirtualVelocity}, we see that the 
two virtual velocities are essentially equivalent. The only difference
is that the online learning rule uses the weighted average velocity 
estimate from the previous time step and the classical rule uses the estimate
from this time step. Given the typical decay time scale of $\lambda$
(which often decays to zero on the order of tenths of a second)
compared to the time scale of $\Delta t$ (on the order of a millisecond),
the difference is negligible.

We can therefore say that the online learning rule, and the classical rule, are
the effectively the same for this simple case.  Using the online learning
formulation, though, makes it easier to leverage more adaptive techniques from
machine learning as discussed in Section~\ref{sec:EquivalenceDiscussion}.

\section{Conclusion and Future Work} \label{sec:conclusion}

In this work, we presented a novel approach to online learning of inverse dynamics modeling errors. Our algorithm directly minimizes a loss function that directly penalizes the error between desired and actual accelerations, enabling the direct execution of raw acceleration policies such as operational space controllers and Linear Quadratic Regulator (LQR) controllers outputting desired accelerations as a function of the robot's state. Using a direct loss overcomes the off-distribution learning issue present in indirect loss approaches which prevail in existing state-of-the art  inverse dynamics learning. We have shown how we can perform online gradient descent on parameters of general nonlinear function approximators to learn an error model of the dynamics. In our extensive evaluations, even with a simple constant error model, by updating and adapting it online, we show that our approach can correct inverse dynamics errors on the fly, for real-world motion generation.

Future work will investigate the use of more complex function approximators, especially friction models and body point forces and torques to compensate for tools, sensors, and other types of loads, as well as interactions between this memory-less adaptive control style learning and learning control style iteration model improvement leveraging these updates.

\section*{Acknowledgements}
This research was supported in part by National
Science Foundation grants IIS-1205249, IIS-1017134,
EECS-0926052, the Office of Naval Research, the Okawa
Foundation, and the Max-Planck-Society. This work was performed
in part while Nathan Ratliff was affiliated with the Max Planck 
Institute for Intelligent Systems.

\small{
\bibliographystyle{plain}
\bibliography{adaptive_control,references}

\begin{thebibliography}{10}

\bibitem{An:1985el}
Chae~H An, Christopher~G Atkeson, and John~M Hollerbach.
\newblock {Estimation of inertial parameters of rigid body links of
  manipulators}.
\newblock In {\em 24th IEEE Conference on Decision and Control}, pages
  990--995, 1985.

\bibitem{An:1988wi}
Chae~H An, Christopher~G Atkeson, and John~M Hollerbach.
\newblock {\em {Model-based Control of a Robot Manipulator}}.
\newblock MIT Press (MA), 1988.

\bibitem{AdaptiveControl2008}
Karl~J. Astrom and Dr.~Bjorn Wittenmark.
\newblock {\em Adaptive Control}.
\newblock Dover, 2nd edition, 2008.

\bibitem{MachineLearningBishop2007}
Christopher~M. Bishop.
\newblock {\em Pattern Recognition and Machine Learning}.
\newblock Springer, 2007.

\bibitem{StochasticLearningBottou2004}
L{\'e}on Bottou.
\newblock {\em Stochastic Learning}, pages 146--168.
\newblock Springer Berlin Heidelberg, Berlin, Heidelberg, 2004.

\bibitem{OptimizationLargeScaleML2016}
L\'{e}on Bottou, Frank~E. Curtis, and Jorge Nocedal.
\newblock Optimization methods for large-scale machine learning.
\newblock 2016.

\bibitem{GeneralizabilityOnlineLearning2004}
N.~Cesa-Bianchi, A.~Conconi, and C.~Gentile.
\newblock On the generalization ability of on-line learning algorithms.
\newblock {\em IEEE Transactions on Information Theory 50}, page 2050–2057,
  2004.

\bibitem{Cesa-BianchiPLG2006}
Nicolo Cesa-Bianchi and Gabor Lugosi.
\newblock {\em Prediction, Learning, and Games}.
\newblock Cambridge University Press, New York, NY, USA, 2006.

\bibitem{craig2005introduction}
John~J Craig.
\newblock {\em Introduction to robotics: mechanics and control}, volume~3.
\newblock Pearson Prentice Hall Upper Saddle River, 2005.

\bibitem{Craig:1987dc}
John~J Craig, Ping Hsu, and S~Shankar Sastry.
\newblock {Adaptive Control of Mechanical Manipulators}.
\newblock {\em The International Journal of Robotics Research}, 1987.

\bibitem{adagrad}
John Duchi, Elad Hazan, and Yoram Singer.
\newblock Adaptive subgradient methods for online learning and stochastic
  optimization.
\newblock {\em Journal of Machine Learning Research}, 12(Jul):2121--2159, 2011.

\bibitem{Gijsberts2013}
Arjan Gijsberts and Giorgio Metta.
\newblock {Real-time model learning using Incremental Sparse Spectrum Gaussian
  Process Regression}.
\newblock {\em Neural Networks}, 41, 2013.

\bibitem{Ijspeert2013}
Auke~Jan Ijspeert, Jun Nakanishi, Heiko Hoffmann, Peter Pastor, and Stefan
  Schaal.
\newblock {Dynamical Movement Primitives: Learning Attractor Models for Motor
  Behaviors.}
\newblock {\em Neural computation}, 25, 2013.

\bibitem{RobustAdaptiveControlIoannou2012}
Petros Ioannou and Jing Sun.
\newblock {\em Robust Adaptive Control}.
\newblock Dover Publications; First Edition, 2012.

\bibitem{adam}
Diederik~P. Kingma and Jimmy Ba.
\newblock Adam: {A} method for stochastic optimization.
\newblock {\em CoRR}, abs/1412.6980, 2014.

\bibitem{Meier2014a}
Franziska Meier, Philipp Hennig, and Stefan Schaal.
\newblock {Incremental Local Gaussian Regression}.
\newblock In {\em Advances in Neural Information Processing Systems 27}. 2014.

\bibitem{Nguyen-Tuong2008a}
D~Nguyen-Tuong, M~Seeger, and J~Peters.
\newblock {Local gaussian process regression for real time online model
  learning and control}.
\newblock {\em Advances in Neural Information Processing Systems}, 22, 2008.

\bibitem{SubgradientMMSCRatliff2007}
Nathan Ratliff, J.~Andrew~(Drew) Bagnell, and Martin Zinkevich.
\newblock (online) subgradient methods for structured prediction.
\newblock In {\em Eleventh International Conference on Artificial Intelligence
  and Statistics (AIStats)}, March 2007.

\bibitem{Righetti_AR_2013}
L.~Righetti, M.~Kalakrishnan, P.~Pastor, J.~Binney, J.~Kelly, R.~Voorhies,
  G.~Sukhatme, and S.~Schaal.
\newblock An autonomous manipulation system based on force control and
  optimization.
\newblock {\em Autonomous Robots}, 2013.

\bibitem{Slotine:1989fd}
Jean-Jacques~E Slotine and Weiping Li.
\newblock {Composite adaptive control of robot manipulators}.
\newblock {\em Automatica}, 1989.

\bibitem{OptimalControlEstimationStengel94}
R.~Stengel.
\newblock {\em Optimal Control and Estimation}.
\newblock Dover, New York, 1994.

\bibitem{ClassicalMechanicsTaylor}
John~R. Taylor.
\newblock {\em Classical Mechanics}.
\newblock University Science Books, 2005.

\bibitem{Vijayakumar2000}
Sethu Vijayakumar and Stefan Schaal.
\newblock {Locally Weighted Projection Regression: An O(n) Algorithm for
  Incremental Real Time Learning in High Dimensional Space}.
\newblock {\em ICML}, 2000.

\bibitem{WassermanAllOfStatistics2010}
Larry Wasserman.
\newblock {\em All of Statistics: A Concise Course in Statistical Inference}.
\newblock Springer Publishing Company, Incorporated, 2010.

\bibitem{zinkevich2003online}
Martin Zinkevich.
\newblock Online convex programming and generalized infinitesimal gradient
  ascent.
\newblock In {\em Proceedings of the 20th International Conference on Machine
  Learning (ICML-03)}, pages 928--936, 2003.

\end{thebibliography}
}

\clearpage
\appendix
\section{More Results}
\label{sec:more_plots}
More result visualizations for the parameter sensitivity analysis of Section~\ref{sec:param_sensitivity_analysis}.

\begin{figure*}[h]
\begin{center}
\includegraphics[height=0.2\textheight]{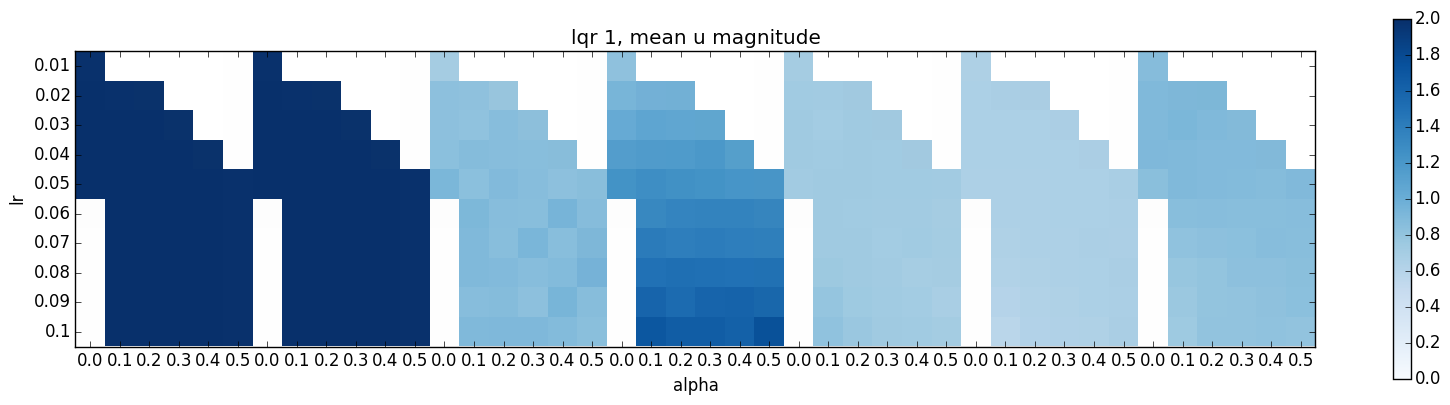}\\
\includegraphics[height=0.2\textheight]{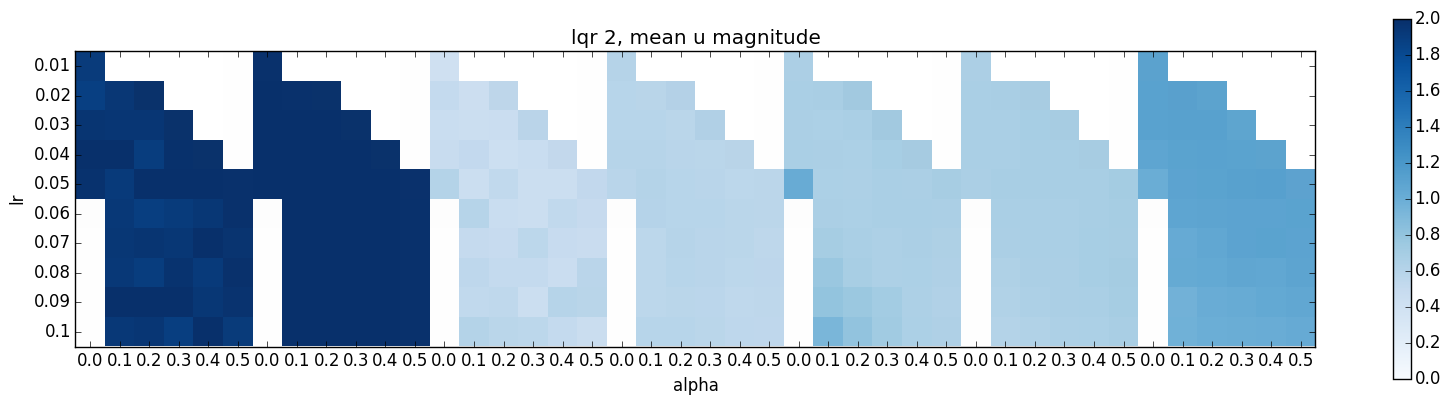}%
\caption{\small Torque magnitudes estimated by the adaptive
  controller $\gamma = 0.95$. (top) LQR \#1, (bottom) LQR \#2. Results are shown  per joint, from left to right joint 1 to 7.}
\label{fig:adaptive_torque_offsets2}
\end{center}
\end{figure*}

\begin{figure*}
\begin{center}
\includegraphics[height=0.2\textheight]{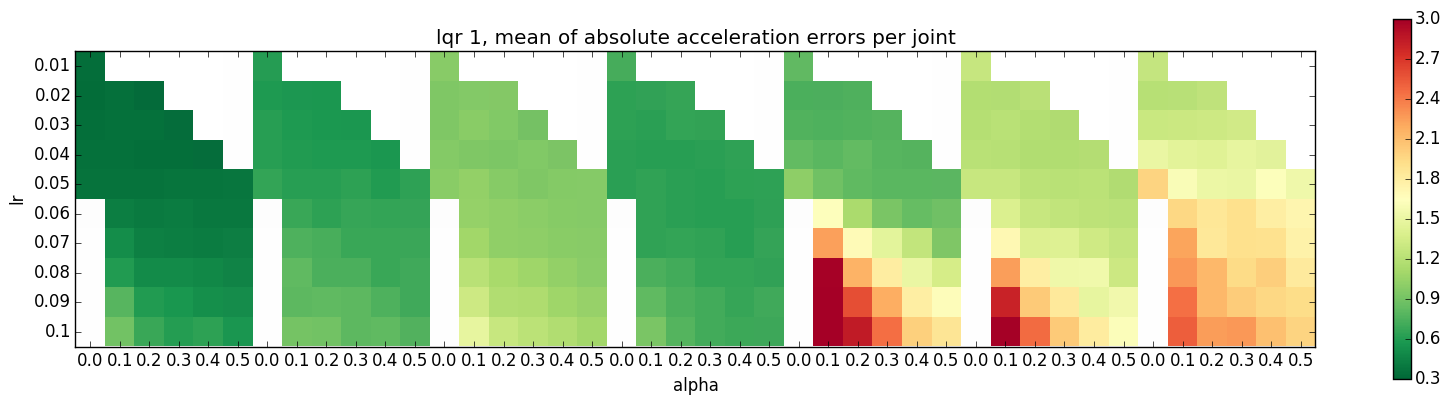}\\
\includegraphics[height=0.2\textheight]{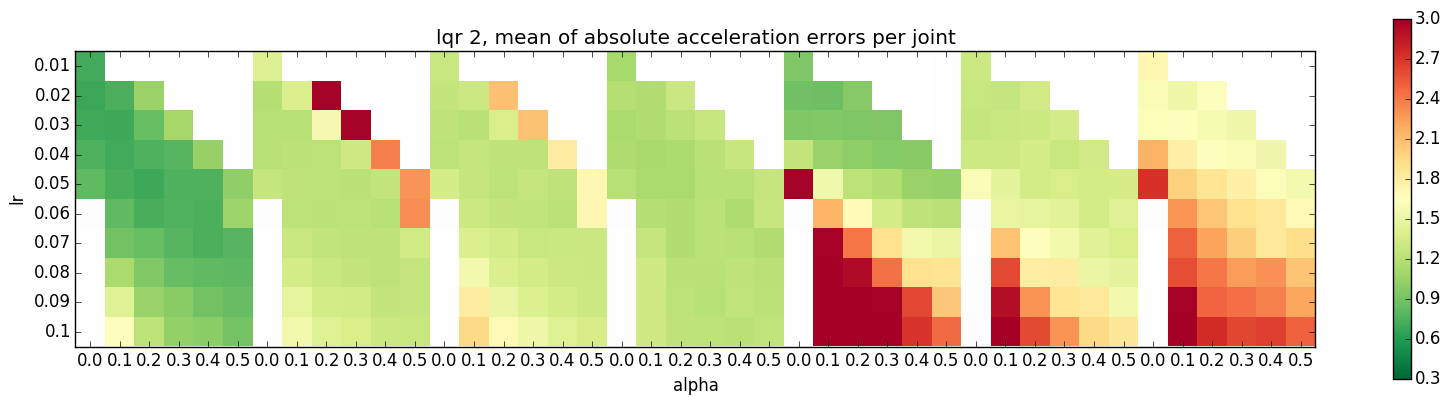}\\
\includegraphics[height=0.2\textheight]{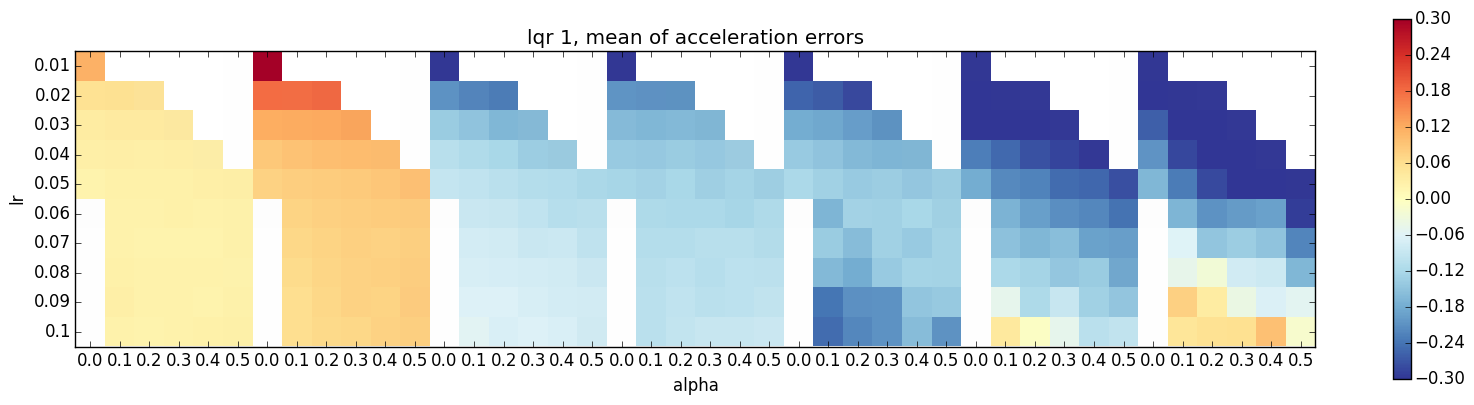}\\
\includegraphics[height=0.2\textheight]{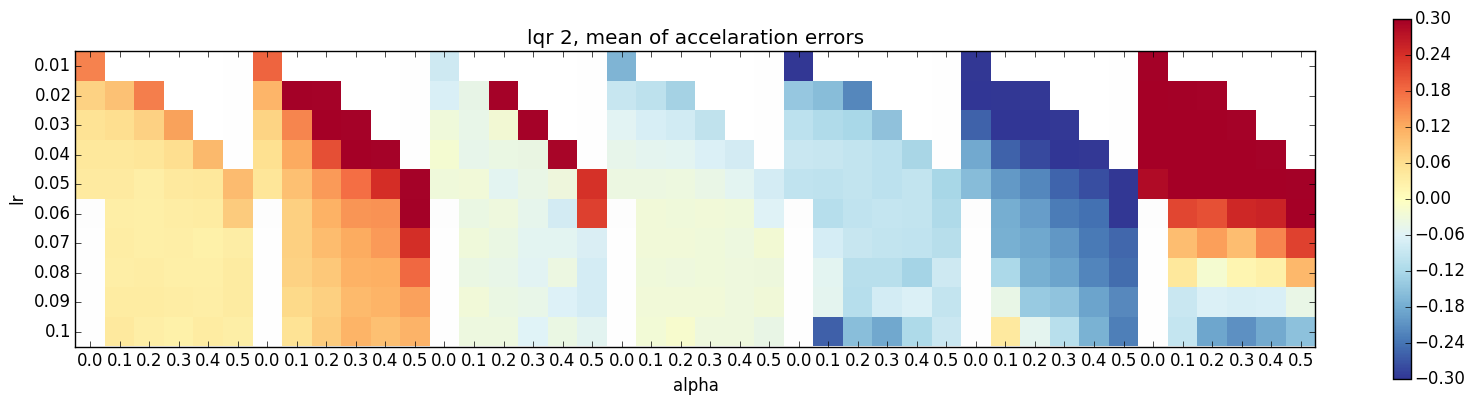}%
\caption{\small The average acceleration error and the error bias for both LQRs with $\gamma = 0.95$. Results are displayed per joint, from left to right joint 1 to 7. From top to bottom: the average absolute acceleration for LQR \#1 and \#2, the mean of the acceleration error bias for both LQRs.}
\label{fig:param_sensitivity2}
\end{center}
\end{figure*}

%
\end{document}